%% file: thesis.tex
\documentclass[a4paper,12pt,times,numbered,print,index]{Classes/PhDThesisPSnPDF}

\usepackage{xcolor}
\usepackage{amsmath}
\usepackage{amssymb}
\usepackage{amsthm}
\usepackage{mathtools}
\usepackage{dsfont}
\usepackage{bm}
\usepackage{textgreek}
\usepackage[absolute,overlay]{textpos}  
\usepackage{xspace}
\usepackage{multirow}
\usepackage{graphicx}
\usepackage{makecell}
\usepackage{cancel}
\usepackage{amsthm}
\usepackage[many]{tcolorbox}
\usepackage{algorithm}
\usepackage{algpseudocode}
\usepackage{nicefrac}   
\usepackage{xspace}
\usepackage{multirow}
\usepackage{multicol}

\input{Preamble/preamble}

\input{thesis-info}

\ifdefineAbstract
\fi

\ifdefineChapter
\fi

\input{definitions.tex}

\begin{document}

\frontmatter

\maketitle

\include{Acknowledgement/acknowledgement}
\include{Abstract/abstract}


\tableofcontents




\printnomenclature

\mainmatter

\part{Probabilistic Reasoning and AI}

\include{Chapter1/chapter1}

\include{Chapter2/chapter2}

\include{Chapter3/chapter3}

\include{ChapterSOTA/chapterSOTA}

\include{Chapter4/chapter4}


\begin{spacing}{0.9}


\bibliographystyle{apalike}
\cleardoublepage
\bibliography{references} 



\end{spacing}







\printthesisindex 

\end{document}

%% file: thesis-info.tex
\title{Probabilistic Circuits as Reasoning Machines in Artificial Intelligence \\ (Part I)}

\author{Robert Peharz}

\dept{Institute of Machine Learning and Neural Computation}

\university{Graz University of Technology}
\crest{\includegraphics[width=0.4\textwidth]{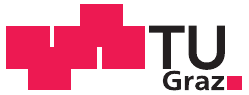}}

\renewcommand{\submissiontext}{Cumulative Habilitation Thesis to obtain the Venia Docendi in}

\degreetitle{Applied Computer Science}



%% file: definitions.tex
\definecolor{robred}{rgb}{.8, 0.0, 0.0}
\definecolor{amber}{rgb}{1.0, 0.49, 0.0}
\definecolor{deepmagenta}{rgb}{0.8, 0.0, 0.8}
\definecolor{robmagenta}{rgb}{0.6, 0.0, 0.6}
\definecolor{robblue}{rgb}{0.0, 0.4, 0.75}
\definecolor{drobblue}{rgb}{0.0, 0.3, 0.77}
\definecolor{robgreen}{rgb}{0.0, 0.5, 0.0}
\definecolor{royalblue(web)}{rgb}{0.25, 0.41, 0.88}
\definecolor{ao(english)}{rgb}{0.0, 0.5, 0.0}
\definecolor{royalazure}{rgb}{0.0, 0.22, 0.66}
\definecolor{ultramarine}{rgb}{0.07, 0.04, 0.56}
\definecolor{gold}{rgb}{1.0, 0.84, 0.0}
\definecolor{darkgold}{rgb}{0.86, 0.73, 0.0}
\definecolor{brickred}{rgb}{.75, 0.32, 0.32}
\definecolor{frorange}{rgb}{.98, 0.59, 0.27}

\newtheorem{definition}{Definition}
\newtheorem{example}{Example}

\DeclareMathAlphabet\mathbfcal{OMS}{cmsy}{b}{n}

\newcommand{\ith}[1]{${#1}^{\text{th}}$}

\newcommand{\halfbline}{\vspace{0.5\baselineskip}\xspace}
\newcommand{\bline}{\vspace{\baselineskip}\xspace}

\newcommand{\graph}{\mathcal{G}}

\newcommand{\outn}{\mathbf{out}}
\newcommand{\inn}{\mathbf{in}}

\newcommand{\XS}{\mathcal{X}}
\newcommand{\XXS}{\mathbfcal{X}}
\newcommand{\X}{\bm{X}}
\newcommand{\x}{\bm{x}}

\newcommand{\YYS}{\mathbfcal{Y}}
\newcommand{\Y}{\bm{Y}}
\newcommand{\y}{\bm{y}}

\newcommand{\Z}{\bm{Z}}
\newcommand{\z}{\bm{z}}

\newcommand{\reals}{\mathbb{R}}
\newcommand{\naturals}{\mathbb{N}}

\newcommand{\E}{\mathds{E}}

\newcommand{\prob}{\mathds{P}}

\newcommand{\p}{p}

\newcommand{\cbar}{\,|\,}
\newcommand{\ci}{\perp\!\!\!\perp}

\newcommand{\node}{\mathsf{N}}
\newcommand{\sumnode}{\mathsf{S}}
\newcommand{\prodnode}{\mathsf{P}}
\newcommand{\distnode}{\mathsf{D}}

\DeclareSymbolFont{mathx}{U}{mathx}{m}{n}
\DeclareMathAccent{\widecheck}    {0}{mathx}{20}

%% file: Acknowledgement/acknowledgement.tex

\begin{acknowledgements}

The people who need to be mentioned foremost are my lovely family, who have patiently endured, sometimes enjoyed, and always supported my academic endeavours:  
Thank you, Elisabeth, Benjamin, Raffael, and Hannah.

This thesis is about probabilistic circuits and tractable probabilistic models.  
We are a small but growing community, continuously striving to promote and inspire enthusiasm for our approach to AI and probabilistic modeling.  
It has been a fun ride, and I hope it continues to be for decades (and centuries) to come.  
It wouldn't be such a great pleasure to work in such an exciting field without the right allies and peers, some of whom are (in alphabetical order):  
Cory Butz,  
Cassio de Campos,  
YooJung Choi,  
Alvaro Correia,
Nicola Di Mauro,  
Floriana Esposito,  
Gennaro Gala,  
Hong Ge,  
Robert Gens,  
Zoubin Ghahramani,  
Kristian Kersting,  
Steven Lang,  
Thomas Liebig,  
Lorenzo Loconte,  
Alejandro Molina,  
Martin Mundt,  
Jhonatan Oliveira,  
Franz Pernkopf, 
Pascal Poupart,
Erik Quaeghebeur,  
Carl Rasmussen,  
Xiaoting Shao,  
Karl Stelzner,  
Pranav Subramani,  
PingLiang Tan,  
Martin Trapp,  
Isabel Valera,  
Guy Van den Broeck,  
Fabrizio Ventola,  
Antonio Vergari,  
Thomas Wedenig,  
and  
Yang Yang.
 
Furthermore, I would like to thank the Austrian Science Fund (FWF) for supporting my work in the context of the Cluster of Excellence ``Bilateral AI'' (10.55776/COE12).

\end{acknowledgements}

\begin{textblock*}{20mm}(92mm,168.77mm) 
\fontsize{1.}{1.5}\selectfont\textcolor{black}{\dots and of course Bugs Bunny}
\end{textblock*}

%% file: Abstract/abstract.tex
\begin{abstract}
This cumulative habilitation thesis studies probabilistic circuits (PCs) as a powerful and tractable framework for reasoning and learning under uncertainty in artificial intelligence (AI).
It first advocates for probability as a core language for AI, emphasizing its connections to logic and information theory; the conceptual simplicity of probabilistic reasoning---based primarily on the sum and product rules; the parallels between probabilistic inference and human cognition; and the role of probability in optimal decision making.
However, probability also faces significant computational challenges, as probabilistic inference is NP-hard in almost all probabilistic models. 
PCs address these challenges through structural constraints that ensure exact computation of a wide range of inference queries in polynomial time, such as marginals, conditionals, most probable explanations, expectations, and more advanced inference tasks.
This thesis synthesizes a decade of research across foundations, algorithmic developments, and empirical validation of PCs.
Key contributions highlighted in this work are foundational theory of PCs, Bayesian approaches for learning PCs, scalable implementations and integration with deep learning, hybrid models that combine PCs with intractable models, and connections with symbolic machine learning paradigms.
\end{abstract}

%% file: Chapter1/chapter1.tex
\chapter{Introduction}

In the last one or two decades we have witnessed a remarkable explosion of the fields of \emph{artificial intelligence} (AI) and \emph{machine learning} (ML).
As research fields, AI and ML have gained tremendous momentum, reflected by an about tenfold increase of papers submitted to leading conferences over the last 10 years.
In the digital industries ML is widely applied, ranging from more ``classic'' applications---such as spam mail detection, optical character recognition, machine translation, computer vision, and recommender systems---to more recent creative applications fuelled by advances in generative modelling, as exemplified by diffusion-based models for images and the prominent ChatGPT system.
More generally, AI and ML are perceived as integral parts of the so-called ``fourth industrial revolution,'' promising a massive acceleration in automation of cognitive tasks and a tight integration of the physical and digital worlds.

\section{What is AI (roughly)?}     
\label{sec:what_is_ai}

While being an active and trendy technology, it is hard to fully specify what AI actually is.
According to ``Artificial Intelligence: A Modern Approach'', the standard textbook by Russell and Norvig \cite{russel2010}, we can organize AI into four categories, spanned by two dichotomies: ``human-oriented vs.~rational AI'' and ``thinking vs.~acting''.
Human-oriented AI strives to imitate abilities of human intelligence, while rational AI formalizes intelligence via some well-principled measure of performance.
While there is overlap between the human-oriented and rational approaches to AI, one can think of rational AIs which are rather distinct from human intelligence, and, vice versa, rational approaches might be insufficient to explain or imitate human intelligence.
Thus, it makes sense to distinct these two styles of AI.
This thesis is located in the ``rational AI'' segment.

The second dichotomy---``thinking vs.~acting''---distinguishes more ``internal'' abilities such as reasoning and cognition from more ``external'' abilities such as acting in an environment and making good decisions.
This distinction also makes sense, even though ``acting'' surely requires ``thinking'' to a certain extent.

While defining AI in a precise way poses some challenges, we can attempt to specify some necessary features or desiderata which any AI system needs to display to a certain degree:
\begin{itemize}
\item \textbf{Knowledge Representation and Modelling.} 
It is a save to assume that any AI needs to have the ability to store information.
More generally, we might assume a certain level of structural organization of information---broadly denoted as \emph{knowledge representation}---facilitating subsequent tasks of the AI system.
Perhaps synonymously to knowledge representation we might also say that an intelligent system needs the ability to represent a \emph{model} of the world in which it is embedded.
In a generous interpretation, knowledge representation and models can take various forms, ranging from explicit and formal representations, such as axiomatic systems, to implicit ones such as information stored in neural network weights.
\item \textbf{Reasoning.} 
Additionally, an AI system will require the ability to \emph{reason}, i.e., to do something with the information stored and process it into ``new'' insights.
The reasoning machinery will in general depend on the type of knowledge representation. 
A classical example is propositional logic, which would represent knowledge as an axiomatic system together with known facts and use deductive reasoning to derive ``new'' facts.
A classical illustration is the propositional knowledge base
\begin{quote}
Socrates is human \\
All humans are mortal
\end{quote}
And the use of the \emph{modus ponens} to conclude that 
\begin{quote}
Socrates is mortal
\end{quote}
But also common (systems of) equations such as 
$
E = mc^2
$
might be interpreted as a representation of knowledge---relating the three quantities $E$, $m$ and $c$---and algebraic manipulation as a reasoning process allowing us, for example, to conclude that ${E \over m} = c^2$. 
\end{itemize}
These two requirements, knowledge representation and reasoning, are arguably necessary features for any AI system.
They are probably not sufficient, as they are also fulfilled on a small scale by rather simple systems such as pocket calculators, which are not commonly perceived as ``intelligent.''
However, some logic-based AI systems in the early area of AI would also be confined to these two aspects, albeit with a high degree of complexity.

Besides knowledge representation and reasoning one would also include the desiderata of \emph{machine learning}, the nowadays most prominent branch of AI:
\begin{itemize}
\item \textbf{Learning.} According to Simon \cite{Simon1983} learning denotes adaptive changes in a system, enabling to perform tasks more effectively and efficiently.
Learning thus includes a certain dynamic element in AI, allowing to automatically modify, extend, or complement the knowledge representation mentioned above.
This does not mean that learning happens necessarily online (over the lifetime of the AI)---in contrary, many machine learning systems nowadays learn in an offline fashion. 
Rather, learning represents a method to incorporate and exploit information which was \emph{not present at design time of the AI}, regardless of whether learning happens online or offline.
\end{itemize}
The desiderata \emph{knowledge representation}, \emph{reasoning}, and \emph{learning} may serve as a minimalist definition of rational AI or, at the very least, describe its necessary features.
Russell and Norvig additionally mention the desiderata \emph{perception} (\emph{computer vision}, \emph{computer audition}), \emph{robotics}, and \emph{natural language processing}. 
In my own opinion, these are natural requirements but are more features describing human-oriented AI.

This habilitation thesis describes the essence of my work on the ``minimalist core notion of rational AI'' described above, putting knowledge representation, learning and reasoning in the center of attention.

\section{The Role of Probability}

A central working hypothesis in this work is to recognize \emph{probability} as a rigorous and consistent tool for \emph{reasoning} and \emph{learning}.
In the following, I briefly summarize arguments for why probability is an excellent core language for AI.

\subsubsection{Simplicity}

Similar as the reasoning tools sketched above, propositional logic and (systems of) equations, probabilistic reasoning is based on some few \emph{simple} mathematical principles.
Arguably, simplicity is key, when it comes to constructing a long-lasting theory of (artificial) intelligence.

Probabilistic reasoning can essentially be reduced to two simple rules, the \emph{sum rule} and the \emph{product rule} \cite{ghahramani2015probabilistic}.
Semantically, the sum rule, also called \emph{marginalization rule}, allows us to \emph{consistently incorporate} unknown facts in our reasoning process, while incorporating the uncertainty associated with these unknown facts.
The semantics of the product rule, also called \emph{conditioning rule}, is to \emph{inject information} in our reasoning process, that is, basing our reasoning on observed facts.
Any probabilistic reasoning task, denoted as \emph{probabilistic inference}, can then be reduced to a well-prescribed cascade of these two rule, allowing us to propagate information from observations to predictions while consistently treating the uncertainties described by our model.

Note that simplicity means here \emph{conceptual} simplicity, not \emph{computational} simplicity.
Indeed, probabilistic inference (applying the sum and product rule) is in general a ``computational nightmare'' as it requires summations or integrations which in general scale exponentially in the problem dimensionality. 
Working around this computational nightmare is a core goal of \emph{probabilistic circuits}, and hence, of this thesis.

\subsubsection{Decision Making}

Probabilistic reasoning is tightly connected with \emph{decision making} under uncertainty.
A classical line of argument are \emph{Dutch book arguments}, which assert that the betting strategy of an agent (human or artificial) must correspond to \emph{some} probability distribution over gambling outcomes.
Otherwise, a combination of bets, a so-called Dutch book, can be constructed which the agent will accept according to its strategy, which, however, will incur guaranteed total loss.
Dutch book arguments do not discuss optimality of the agent's betting strategy, but rather are a sanity check, arguing which kind of strategies are in principle admissible, namely, those based on probabilistic reasoning.

More generally, probability lies at the heart of \emph{decision theory} \cite{berger1990statistical} which defines the notion of \emph{optimal decision} in mere probabilistic terms.
Specifically, an optimal decision regarding some target quantity $Y$ is defined as the minimizer of the \emph{expected loss}, that is, the expectation of a some loss function $\ell(y', y)$, expressing the regret when predicting $y'$ while the true value of $Y$ is $y$.
In that sense, knowledge of the true underlying joint distribution together with rigorous probabilistic inference inevitably leads to optimal decision, in the sense of minimal expected (or long-term) regret under this distribution.

\subsubsection{Qualitative Similarities to Common Sense Reasoning}

Furthermore, probabilistic reasoning shares many qualitative features with (human) common sense reasoning, as discussed by Pearl in his classical book ``Probabilistic reasoning in intelligent systems: networks of plausible inference'' \citep{pearl1988probabilistic}.

First, probabilistic reasoning features \emph{non-monotonic reasoning}, i.e.~the effect that the degree of plausibility is not a monotonic function of the amount of available information.
For example, common sense judges the statement ``Tim can fly'' as of low plausibility, but learning the information that ``Tim is a bird'' would lead to an increase of this plausibility.
If, however, one learns the additional piece of information that ``Tim is sick'' one would consequently lower this plausibility again.
This mechanism of \emph{non-monotonic belief update}, i.e.~that the plausibility of a statement is not monotonously increasing or decreasing with the amount of information available, is naturally reflected in probabilistic reasoning.
Specifically, \emph{conditioning} (the product rule) displays this feature, as conditional distributions might be arbitrarily different depending on the conditioned information.
Several classical AI systems for reasoning under uncertainty were lacking the ability of non-monotonous reasoning \cite{pearl1988probabilistic}.

Another canonical example where probabilistic reasoning parallels human cognitive reasoning is the \emph{explaining away phenomenon}: two a-priori unrelated causes become correlated upon observing a common effect. 
For example, a PhD student in ML might observe that ($A$) their experiments deliver dissatisfying results.
Two reasons for this might be ($B$) the whole theory and foundations of the work are flawed, or ($C$) there is a bug in the implementation. 
A priori, statements $B$ and $C$ can be assumed to be uncorrelated. 
However, when $A$ is observed, then common sense reasoning suggests that learning $C$ makes $B$ less plausible (bringing some relief to the PhD student), showcasing that $B$ and $C$ have become correlated upon observing $A$.
The explaining away effect, as well as other common sense patterns determining \emph{relevance} among quantities, is well reflected in probabilistic reasoning \cite{pearl1988probabilistic}.


\subsubsection{Bayesianism and Consistency with Logic}

Unlike the classical ``hard'' reasoning tools, such as logic and systems of equations, probability adds a notion of \emph{uncertainty} which is a pressing requirement imposed by the real world.
A nice feature of probability is that, when restricted to the extreme probabilities $0$ and $1$, probability emulates classical propositional logic.
Moreover, if one accepts the assumptions of \emph{Cox's theorem}, then probability is the \emph{only} sound extension of classical logic to \emph{continuous} \emph{truth} or \emph{plausibility} values, 
so that one is de facto forced to accept probability in order to be compatible with logic \cite{jaynes2003probability}.

Cox's theorem is often cited as a theoretical underpinning of the \emph{Bayesian approach},
which naturally unifies reasoning and learning.
The Bayesian perspective allows us to put probabilities on essentially any uncertain quantities of interest, not only those which can arguably be repeated, as required by the classical frequentist notion of probability.\footnote{In a nutshell, the classical dispute between Bayesians and frequentists is about how the notion of probability shall be ``correctly interpreted.'' 
}
Probabilistic modelling is then but the task of constructing a joint distribution expressing the dependencies and uncertainties between all involved model parameters\footnote{Here, parameter can mean essentially any unknown quantity of interest, not merely model weights.} and the data.
The typical modelling approach is to specify a \emph{prior distribution} over parameters and a \emph{likelihood} relating parameters with the data.
The division into prior and likelihood, however, is usually a rather arbitrary choice \cite{ghahramani2013bayesian}. 
Then, in the Bayesian approach, learning is but probabilistic  reasoning (posterior inference) about parameters given data.

\subsubsection{Information Theory}

Information theory was originally designed as a tool of communication engineering \cite{shannon1948mathematical}, but quickly found applications in many other disciplines, such as biology, psychology, and fundamental physics \cite{jaynes1957information}.
The basic insight is to quantify the information content of an event by the \emph{negative logarithm of its probability}.
As information content and probabilities are merely non-linear but monotonous transformations (log vs.~exp) of each other, they are equivalent descriptions of the same quantity.
This equivalence is satisfying, as it indicates that (logs of) probabilities are merely measuring information, or actually lack thereof, serving as a natural description of \emph{uncertainty}.

Here, the reason \emph{why} the information is missing is a subordinate matter and one can avoid vague talk about ``true'' randomness.
Otherwise, one would need to worry whether an opponent’s poker cards are truly random (they are not for them, and I could simply grab the cards and check myself), or the outcome of a game of roulette (which, with sufficiently accurate physical measurement could be predicted arbitrarily well), or the pseudo-random behaviour of a chaotic system which seems random but is actually perfectly deterministic, or the effects on the level of particle physics (Heisenberg's uncertainty principle introduces a fundamental limit for physical beings).
While in these examples that nature of ``randomness'' seems quite different, this does not matter for modelling and reasoning purposes---one simply quantifies uncertainty, or lack of information, with (log) probabilities.
As the information state is generally agent-dependent,
this line of argument is reinforcing a Bayesian approach to probabilistic reasoning \cite{mackay2003information}.

\vspace{\baselineskip}

This thesis puts forward the working hypothesis that probability is an excellent core language for AI, due to its conceptual simplicity (and hence the possibility to be automated), its ability to express and process uncertainty, as well as many compelling connections to logic, human reasoning, decision theory and information theory.

\section{Why Probabilistic Circuits?}

While probability is arguably an excellent learning and reasoning language for AI, it poses substantial computational challenges.
For example, \emph{probabilistic graphical models} (PGM) \cite{koller2009probabilistic}, a prominent framework for probabilistic AI, are decorated with many hardness results concerning inference and learning.
Often, it is relatively easy to construct a PGM accurately reflecting the domain of interest, but in which performing sound probabilistic reasoning (via the sum and product rule) has exponential computational cost in the so-called \emph{tree-width} of the PGM.
Roughly speaking, PGMs and other types of probabilistic models might be easy to represent but ``hard to run''.


This is in stark contrast to, say, neural networks which are constructed in a way that inference remains easy.
Of course, the shortcut here is that neural networks do not perform fully fledged probabilistic reasoning, but only learn a single (or few) reasoning task(s).
For example, if one learns a neural classifier predicting class label $Y$ from a feature vector $\X$ by minimizing cross-entropy, one really just estimates the conditional distribution $p_{Y|\X}$ with maximum likelihood, i.e.~minimizing the Kullback-Leibler divergence between the model distribution and the empirical data distribution.
The very same network, however, would have trouble to predict a portion of the features $\X' \subset \X$ given the rest of the features $\X'' = \X \setminus \X'$ while ignoring the label $Y$ altogether. 
Such a predictor cannot be derived in a meaningful way from $p_{Y|\X}$.
Rather, one requires a model over the \emph{whole joint} $p_{Y,\X}$, which would allow to compute the desired conditional distribution $p_{\X' | \X''}$.
Of course, there are ample neural networks representations for whole joint distributions, generally known as \emph{(deep) generative models} \cite{tomczak2021deep}.
However, similar as for PGMs, probabilistic inference is generally hard in these models.

Can we get best of both worlds?
That is, can we design models that---like PGMs and other probabilistic models---represent \emph{full} joint distributions over the domain of interest, while at the same time allowing to perform flexible inference with the same ease and computational complexity as neural networks?  
The answer is \textbf{yes, with probabilistic circuits (PCs)!}
One way to look at PCs is that they are a kind of hybrid model between PGMs and neural networks.
Like PGMs---and unlike ``standard'' neural nets---they are representations of joint distributions. 
Unlike PGMs they are able to perform \textbf{exact} and \textbf{efficient} inference, that is, they compute the sum and product rule---as well as many other inference queries---in essentially the same manner and with the same computational cost as neural networks, i.e.~by performing one or a few network passes.

The key term here is \textbf{structure}:
While PCs can be interpreted as neural networks, they need to obey particular constraints guaranteeing that the desired reasoning operations remain tractable.
Interestingly, these structural constraints appear at many places in literature, within the probabilistic modelling setting but also in related disciplines such as logic \cite{darwiche2002knowledge} and databases \cite{olteanu2016factorized}.
Within the probabilistic reasoning domain, these structural patterns appear over and over again, be it in exact inference algorithms such as \emph{variable elimination} \cite{darwiche2009modeling}, the \emph{junction tree algorithm} \cite{koller2009probabilistic}, and \emph{and/or search spaces} \cite{dechter2007and} or tractable probabilistic models such as \emph{arithmetic circuits} \cite{darwiche2003differential}, \emph{sum-product networks} \cite{poon2011sum}, \emph{cutset networks} \cite{rahman2014cutset}, and \emph{probabilistic sentential decision diagrams} \cite{kisa2014probabilistic}.
Really, PCs do not represent a novel model class.
Rather, they aim to consolidate these existing lines of research under one umbrella and using one syntax \cite{vergari2020probabilistic,choi2020probabilistic} and systematically delineate their differences in terms of structural constraints and corresponding tractable reasoning routines~\cite{vergari2021compositional}.
In a nutshell, PCs are a proposed \textbf{lingua franca for tractable probabilistic modelling}.

Evidently, PCs don’t offer a free lunch nor do they possess a magical secret sauce to make hard probabilistic inference problems all of the sudden easy. 
At their core, PCs are models too and serve as an approximations of reality. 
Their structural constraints ensure that the primary goal---probabilistic inference---remains tractable, but this comes at the cost of limiting the choice of neural network structures. 
These constraints naturally impose limits on the expressive efficiency of PCs, meaning that some distributions, which can be compactly described as an intractable model, may require exponentially larger representations as a PC. 
Therefore, the key strategy in PCs is to find the best possible approximation of the data distribution under the constraint that inference remains tractable.

This cumulative habilitation thesis describes my contributions to the field of PCs over the last ten years and is organized as follows:
In Chapter~\ref{chap:probability}, I introduce basic concepts of probability theory and probabilistic reasoning. 
In Chapter~\ref{chap:probabilistic_reasoning}, I illustrate the idea of probability as elegant and widely applicable reasoning tool on the level of distribution functions.
Chapter~\ref{chap:SOTA} gives a concise overview of state-of-the-art in probabilistic modeling.  
In Chapter~\ref{chap:probabilistic_circuits}, I introduce the framework of PCs and illustrate the contributions I made to this research field.
The relevant publications are listed in Part~II.

%% file: Chapter2/chapter2.tex
\chapter{Basic Probability Theory}
\label{chap:probability}

The basic concept in probability is the \emph{probability space}.
The definition of probability space is somewhat abstract and practitioners like to quickly forget it as soon as they get exposed to more intuitive notions like \emph{random variables} and \emph{probability densities}.
However, it pays off to embrace such foundational concepts to a certain extent.
Probability as a mathematical discipline is relatively young, only dating back about 300-400 years, and a firm and concise description of probability was not immediate.
Today we have arrived at a compact and rigorous formalisation of probability---the \emph{probability space}---which fits on roughly half a page and accurately describes the nature of probability.

The probability space serves as an insurance and contract:
probabilistic methods and models can get fairly abstract and complex and one might get confused about whether what one is doing is still sound.
The rule of the game is: can we reduce it back to \emph{some} probability space (or at least prove its existence)? 
If yes, we can call it probability.
If not, it should be called something else.
In other words, the probability space is the ``Linux kernel'' of probabilistic AI: we usually don't need to work with it, but knowing how it works will make us better Linux users.

In this chapter I just introduce the rudimentary ideas of measure-theoretic probability theory, in order to establish its role as central reasoning instrument in AI.
A rigorous treatment can be found for example in \cite{billingsley1995,rosenthal2006first}.

\section{Measurable Space}

The probability space is based on a structured set denoted as \emph{measurable space}.

\begin{definition}[Measurable Space]    \label{def:measurable_space}
Let $\Omega$ be any \emph{non-empty set} and $\Sigma$ a \emph{sigma-algebra} over $\Omega$, i.e.~$\Sigma$ is a set of sub-sets $A \subseteq \Omega$.
In the following, let $\bar A \coloneqq \Omega \setminus A$ denote the \emph{complementary set} of $A$.
The sigma-algebra needs to satisfy the following properties:
\begin{itemize}
\item $\Omega \in \Sigma$
\item $A \in \Sigma$ implies that $\bar A \in \Sigma$   \hfill (\emph{closed under complement})
\item for any countable collection of events $\{A_i\}_i$ it holds that $\bigcup_i A_i \in \Sigma$. \\ \mbox{}  \hfill (\emph{closed under countable union})
\end{itemize} 
The tuple $(\Omega, \Sigma)$ is a \emph{measurable space}.
\end{definition}

Basically, a measurable space is just a non-empty set ($\Omega$) plus some structure represented as a collection of sub-sets ($\Sigma$).
The set $\Omega$ is called the \emph{sample space} and the elements $\omega \in \Omega$ are called \emph{atomic events}.
The basic picture in probability is that an element of $\omega$ is selected \emph{at random} from $\Omega$, where ``at random'' really just means that we are lacking the information which $\omega$ is selected. 
The basic idea of probability is to mathematically model and treat this lack of information.

It is important to note that in general we should \emph{not} assume that an $\omega$ is selected \emph{repeatedly} from $\Omega$, as it is assumed in many classical statistics texts, but rather that \emph{exactly one} $\omega$ is selected, and we do not know which one.
The point here is that assuming repeated selection is unnecessarily limiting the theory, and assuming that exactly one $\omega$ is selected is in fact the more general picture.
Repeated selection of $\omega$ is easily subsumed, by ``copying the sample space'', i.e.~one defines $\Omega_i = \Omega$ for $i = \{1, 2, \dots\}$ and constructs a new sample space as the Cartesian product of the copies $\Omega' \coloneqq \bigtimes_{i} \Omega_i$.
In this duplicated sample space $\Omega'$, the atomic events are now sequences of draws $\omega' = (\omega_1, \omega_2, \dots)$.
This construction is also valid for infinite sequences, i.e.~when $i$ ranges over the entire $\naturals$.

Besides the sample space $\Omega$, the probability space contains a sigma-algebra $\Sigma$, a system of subsets $A$ of $\Omega$, which needs to obey the three rules in Definition \ref{def:measurable_space}.  
It follows automatically from these rules, that also (i) $\emptyset \in \Sigma$ and (ii) any countable intersection $\bigcap_{i} A_i$ of $A_i \in \Sigma$ is again in $\Sigma$, i.e.~the sigma-algebra is also \emph{closed under countable intersection}.

\subsection{Events are Sets}

The elements of $\Sigma$ are called \emph{events}.
This nomenclature hints that while the atomic events $\omega$ represent the ``finest resolution'' of the universe $\Omega$, the sigma-algebra $\Sigma$ contains events at a more ``macroscopic level''.
Example~\ref{ex:events} illustrates this concept.

\begin{tcolorbox}[width=\textwidth,]
\begin{example}[Events for a die throw]
\label{ex:events}
When modelling a die, a suitable sample space is $\Omega = \{1,2,3,4,5,6\}$ representing all six possible outcomes of the die throw. 
We might decide that this is too fine-grained for our purposes and that---for whatever reason---we are just interested in two events, namely (i) whether the die shows a \emph{prime number} and (ii) whether it shows an \emph{odd number}.

The die shows a prime number if it turns out that $\omega$ is contained in the set $A_p = \{2,3,5\}$ and it shows an odd number if $\omega$ is in $A_o = \{1,3,5\}$.
One can convince oneself that---as soon as the sample space $\Omega$ is fixed---indeed any thinkable event can be encoded by whether or not $\omega \in A$ where $A$ is some subset of $\Omega$.

We can find \emph{the most concise sigma-algebra} containing $A_p$ and $A_o$ by starting with \mbox{$\Sigma_0 = \{A_p, A_o, \Omega\}$} and iterating 
\begin{align*}
\Sigma_{i+1} & \leftarrow \Sigma_i \cup \{\bar A \colon A \in \Sigma_i\} \cup \{ A \cup B \colon A, B \in \Sigma_i\}
\end{align*}
that is, we repeatedly include all complements and pairwise unions of events.
This process will---for any finite $\Omega$---converge and deliver the most concise sigma-algebra containing the desired events.
Here we get  
\begin{align*}
\Sigma = 
& \{\emptyset, 
\{1\}, \{2\},  
\{1,2\}, \{3,5\}, \{4,6\},
\\
& \{1,3,5\}, \{1,4,6\}, \{2,3,5\}, \{2,4,6\}, \\ 
& \{1,2,3,5\}, \{1, 2, 4, 6\}, \{3,4,5,6\}, \\ 
& \{1,3,4,5,6\}, \{2,3,4,5,6\}, \Omega\}
\end{align*}
We see that this sigma-algebra does not contain all of the singleton atomic events---it contains $\{1\}$ and $\{2\}$, but for example $\{3\}$ and $\{4\}$ are not contained in $\Sigma$.
That is, $\Sigma$ represents a more ``macroscopic'' view of $\Omega$ and contains only $A_p$ and $A_o$ plus all the necessary events to get a valid sigma-algebra.
\end{example}
\end{tcolorbox}

After fixing $\Omega$, any thinkable event can be described as whether $\omega$ is contained in some prescribed set $A \subseteq \Omega$ and hence any such subset $A$ deserves the name ``event''.  
Thus, we will sometimes say things like ``event $A$ happens,'' meaning really that $\omega \in A$.
The basic role of $\Sigma$ is to collect the events we are interested in.
Moreover, the structure of the sigma-algebra just ensures that we have a \textbf{closed logical system} \cite{jaynes2003probability}:
\begin{itemize}
\item The sigma-algebra needs to be closed under complement, which means that for any event we consider, we also must consider the opposite, i.e., that the event does \emph{not} happen.
This simply amounts to \textbf{logical negation}.
\item The sigma-algebra needs to be closed under countable union, i.e., for any countable collection of events we need also to consider the event that any of them happens.
This amounts to \textbf{logical or}.
\item It follows automatically due to De Morgan's laws that the sigma-algebra is closed under \emph{countable intersection}, i.e., for any countable collection of events we need also to consider the event that all of them happen.
This amounts to \textbf{logical and}.
\item Further logical connectives follow, such as closedness under symmetric set difference (\textbf{logical xor}).
\end{itemize}   
We can now define the notion of probability space.

\section{Probability Space}

While a measurable space is a non-empty set $\Omega$ together with a logical structure $\Sigma$ of sub-sets, a probability space adds the \emph{probability measure}, a function defined on $\Sigma$.
\begin{definition}[Probability Space]    \label{def:probability_space}
Let $(\Omega, \Sigma)$ be a measurable space.
Let $\prob$ be a function $\Sigma \mapsto [0,1]$ mapping events to real numbers between $0$ and $1$, with the properties
\begin{itemize}
\item $\prob(\Omega) = 1$
\item for any countable collection of \emph{disjoint} events $\{A_i\}_i$ it holds that 
$\prob \left( \bigcup_i A_i \right) = \sum_i \prob(A_i)$ \\ \mbox{~}  \hfill \emph{(countable additivity)}
\end{itemize}
Such function $\prob$ is called a \emph{probability measure} defined on $(\Omega, \Sigma)$.
The triplet $(\Omega, \Sigma, \prob)$ is called a \emph{probability space}.
\end{definition}
Thus, a probability space equips a measurable space with a function $\prob$ which assigns to each event a real number---the \emph{probability of the event}---such that the whole $\Omega$ gets assigned $1$ and that probabilities of disjoint events add up. 
Note that from this definition it follows that $\prob(\emptyset) = 0$ and $\prob(\bar A) = 1- \prob(A)$.

The probability space according to Definition \ref{def:probability_space} fully characterizes the nature of probability as understood by current mainstream mathematics.
Any probabilistic argument can be reduced to a sample space $\Omega$, a sigma-algebra $\Sigma$ over $\Omega$ and a probability measure $\prob$ defined on $\Sigma$. 
Given that probability is a general and powerful tool, this definition is a truly compressed description of the matter.

\subsection{Why Sigma-Algebras?}

Newcomers to probability often wonder why sigma-algebras are included in the basic definition of probability space.
Ultimately, we aim to assign probabilities to events, which is the role of the probability measure.
We do we need $\Sigma$? 
Why can't we simply just consider \emph{all} possible events, i.e.~\emph{all} possible subsets $A \subseteq \Omega$? 
In other words, one would want to---without loss of generality---define $\prob$ on the \emph{power set} (set of all subsets):\footnote{Note that the power set is indeed a sigma-algebra, namely the largest possible one.} 
$$
2^\Omega \coloneqq \{A \colon A \subseteq \Omega\}.
$$
In that way, one could omit sigma-algebras altogether and make the basic definition of probability space even simpler.
This would surely be the way to proceed if it was mathematically meaningful.
Sigma-algebras, if one would want them to express a ``coarse-grained'' structure, could be included at a later stage, but not in the core definition of probability.
And indeed, we can safely define $\prob$ on all events, as long as $\Omega$ is \emph{countable}, i.e., \emph{finite} or of the same cardinality as the \emph{natural numbers}.

However, it turns out that when $\Omega$ is uncountable, e.g.~the real numbers $\Omega = \reals$, the power set becomes ``too big'' to be used as sigma-algebra.
Even something simple such as a uniform distribution over some interval of the real line cannot be constructed in a consistent way.
These troubles are not private to probability theory but go back to measure theory, which is concerned with assigning \emph{measures} such as volume, length, mass---and also probability---to sets.
The classical question of measure theory was whether the notion of \emph{length} can be generalized from intervals such as $[a,b]$, which naturally has length $b-a$, to arbitrary subsets on the real line.
The answer to this is negative, meaning that there exist sets which cannot be assigned a meaningful length, as it would break one or more desiderata about such length measure (countable additivity, translation invariance etc.). 
Those non-measurable sets are rather abstract and are constructed in a rather indirect way by employing the \emph{axiom of choice},\footnote{The existence of non-measurable sets is in fact equivalent to the axiom of choice.} 
An easy to follow illustration of the problem is given in Rosenthal's book \cite{rosenthal2006first}.

These \emph{non-measurable} sets should be excluded from the discussion and only \emph{measurable sets} should be included.
Exactly this is the primary usage of sigma-algebras and the reason why they are included in the basic definition of probability.

There is a standard choice of sigma-algebra when one is working with the real numbers, i.e., $\Omega = \reals$, or the $D$-dimensional Euclidean space, i.e.~$\Omega = \reals^D$, namely the \emph{Borel sets}.
The Borel sets include any subset one could (constructively) think of (and many more), forming the \emph{Borel sigma-algebra}.
They are easiest defined via \emph{generated sigma-algebras}.

\begin{definition}[Generated Sigma-Algebra]
\label{def:generated_sigma_algebra}
Let $E$ be any set of subsets of $\Omega$.
The \emph{sigma-algebra generated} by $E$ is 
$$
\Sigma(E) \coloneqq \bigcap_{E \subseteq \Sigma} \Sigma,
$$
that is, the intersection of all sigma-algebras containing $E$.
\end{definition}

Note that in Definition \ref{def:generated_sigma_algebra} we do not require any special properties from $E$, but any collection of subsets of $\Omega$ is fine.
It is quite easy to prove that $\Sigma(E)$ is indeed a sigma-algebra.
Furthermore, we can rightfully call it the smallest possible sigma-algebra containing $E$, since it it contains exactly the sets which need to occur in \emph{all} sigma-algebras containing $E$.

With this tool at hand, the Borel sigma-algebra can be defined as
\begin{definition}[Borel Sets, Borel Sigma-Algebra]
\label{def:borelsets}
The \emph{Borel sigma-algebra} on $\reals$ is defined as 
$$
\mathcal{B} \coloneqq \Sigma(\{ [a,b] \cbar a,b \in \reals, a \leq b\})
$$
that is, the sigma-algebra generated by the set of all closed intervals.
Alternatively, one can use the open intervals, half-open intervals, etc., which are all equivalent definitions of $\mathcal{B}$.
\end{definition}

A likewise definition can be made for $\reals^D$, by using the sigma-algebra generated by (open or closed) hypercubes, or alternatively the set of all open (or closed) balls, etc.

In conclusion of this section: sigma-algebras contain those events one wants to consider and make sure that events are closed under set operations (logic connectives). 
The main reason that they are in the basic definition of probability space is to ``fix a bug'' and make the math work.

\section{Interpretations of Probability}

The definition of probability space requires about half a page and precisely sets the formalism and rules to work with probability. 
While it is a crisp and minimalist description of the nature of probability, it lacks intuition about ``what probability actually is''.
We might think about probability in the following ways.
\begin{itemize}
\item \textbf{Relative Frequencies.}
Relative frequencies are helpful for interpreting probabilities and follow from the \emph{law of large numbers}.
Classically, they served as definition of probabilities, before their formalization as in Definition~\ref{def:probability_space}.

Given a probability space $(\Omega, \Sigma, \prob)$, we might construct $n$ independent copies of it.
Intuitively, when figuring the probability space as an urn containing various coloured balls, we would construct $n$ identical urns and let $n$ independent parties perform the draws.
The result is then a sequence (or set) of independent and identically distributed draws $\bm{\omega} = (\omega_1, \omega_2, \dots, \omega_n) \in \Omega^n$.
The law of large numbers now guarantees that given any event $A \in \Sigma$
$$
\lim_{n \to \infty} ~~ {\sum_{i=1}^n [\omega_i \in A] \over n} \overset{a.s.}{=} \prob(A),
$$  
where $[\cdot]$ denotes the Iverson bracket, evaluating to $1$ for true arguments and $0$ for false arguments.
Hence, $\sum_{i=1}^n [\omega_i \in A]$ is the number of times the atomic event has been picked from $A$ across the independent copies of the probability space.
Normalizing by $n$ gives us a relative frequency between $0$ and $1$.
The abbreviation $a.s.$ means \emph{almost surely}, that is, the equality holds with probability $1$ (in the constructed probability space over sequences $\bm{\omega}$).
Hence, drawing the atomic element repeatedly, independently and in identical manner, the relative frequency of how often event $A$ happens approaches $\prob(A)$.
%
%
\item \textbf{Relaxation of Logic Truth Values.} A probability measure assigns a real number to each of the considered events.
At least for two of these real numbers we have a concrete interpretation, namely for $0$ and $1$.
The whole sample space $\Omega$, but perhaps also other events $A$, get assigned probability $1$, which is naturally interpreted as the logical value $true$.
Conversely, the empty set, but perhaps also other events, get the value $0$, naturally interpreted as $false$.
No matter which $\omega$ is picked, it will certainly be contained in $\Omega$ and it will certainly not be in $\emptyset$.
Moreover, if we restricted the set of return values of $\prob$ to $0$ and $1$, any propositional logic theory can be emulated by constructing a suitable probability space.

In the general case, where probabilities are between $0$ and $1$, we might naturally interpret probabilities as \emph{plausibility values}.
The closer a probability is to $1$ the more plausible is its occurrence and the closer it is to $0$ the less plausible its occurrence.
The rules of probability further dictate that the probability of a union of disjoint events (amounting to a disjunction of logically incompatible statements) should simply add up, which is reasonable.
Furthermore, Cox's theorem \citep{jaynes2003probability} asserts that---under particular ``common sense assumptions''---probability is de facto the only sound extension of propositional logic to real numbers. 
\item \textbf{Information.} 
Information theory provides a natural interpretation of probability as information content.
Shannon \cite{shannon1948mathematical} defined the \emph{information content} of an event $A$ as 
$$-\log \prob(A)$$
based on the desiderata that 
\begin{itemize}
\item an event which occurs certainly should convey zero information
\item less probable events (i.e.~events which are ``closer'' to impossibility $\emptyset$) should convey more information
\item information of \emph{independent events}\footnote{The notion of independence will be defined in Section~\ref{sec:conditional_independence}.} should add up
\end{itemize}
The negative logarithm is the only function satisfying these desiderata, where a degree of freedom concerning the base of the logarithm remains.
When using the logarithm to base $2$ information content is measured in \emph{bits}.
A probability of $0.5$, for example the outcome of a fair coin flip, translates to $1 \, bit$, making intuitively sense.
The logarithm to base $e$ expresses information in \emph{nats}.
As information content of an event is a bijective function of its probability, probability and information content are basically the same thing.
\end{itemize}

None of these interpretations should be treated as the ``ultimate correct one''. 
The question how to ``correctly'' interpret probabilities has been long a point of dispute between ``frequentists'' and ``Bayesians''. 
For frequentists, the interpretation as relative frequencies is more appealing, while for Bayesians the interpretation as plausibility values of information content is natural.\footnote{The Bayesian camp sometimes called probabilities ``subjective beliefs'', which is actually an invitation for being criticised for ``not being objective'' or ``believing instead of knowing''. However, the idea of this name bears wisdom: probabilities represent the information status about a particular domain, which, of course, is always depending on the agent (e.g.~I know my cards in poker, but not yours, and vice versa). To reflect this idea, I would suggest the term ``agent-dependent information status'' rather than ``subjective beliefs''.} 
%
%
It is advisable to move beyond discussions about the ``true nature'' of probability.
Different interpretations of probability---and the ability to switch between them---should be understood as a sign of richness and justification of the theory.  
Regardless of interpretation, we shall embrace probability as what it is at its core: a consistent and rigorous reasoning calculus under uncertainty, based on a few simple principles.

\section{Fundamental Laws of Probability}
\label{sec:probabilistic_reasoning}

\emph{Probabilistic modelling} entails constructing a probability space that represents the domain of interest together with any uncertainties, addressing the desideratum of \emph{knowledge representation} in AI, as discussed in Section \ref{sec:what_is_ai}.
The second desideratum is \emph{reasoning}, i.e.~using our knowledge to answer queries of interest.
An important requirement of any reasoning tool is \emph{conceptual simplicity}, that is, reasoning primitives should be relatively few and easy to understand.
As we will see in Chapter~\ref{chap:probabilistic_reasoning}, the core of probabilistic reasoning relies on a few core reasoning primitives, in particular the \emph{sum rule} and \emph{product rule}.
We will discuss these rules on the level of \emph{distribution functions} (introduced in Section~\ref{sec:distribution_functions}), but they correspond to more fundamental laws on the level of probability measures.
We will briefly discuss these laws in the following.

\subsection{Law of Total Probability (The Sum Rule)}
\label{sec:law_total_probability}

Let a probability space $(\Omega, \Sigma, \prob)$ be given and let $B \in \Sigma$ be an event.
Let $A_1, A_2, \dots$ be a countable collection of \emph{exclusive} (i.e.~$A_i \cap A_j = \emptyset$ for $i\not=j$) and \emph{exhaustive} (i.e.~$\cup_i A_i = \Omega$) events, that is, they form a partition of $\Omega$.
Let now
$
B \cap A_i
$
be the \emph{joint event} that both $B$ and $A_i$ happen.
The \emph{law of total probability} states
\begin{equation}
\prob(B) = \sum_i \prob(B \cap A_i).
\end{equation}
This law follows directly from the countable additivity property of probability measures (Definition \ref{def:probability_space}) since the joint events $B \cap A_i$ form a partition of $B$, hence the sum of their probabilities must equal $\prob(B)$.

\begin{figure}
\begin{tcolorbox}[width=\textwidth,]
\begin{example}[Law of Total Probability]
\label{ex:total_probability}
Consider a sample space $\Omega$ representing a population of people, i.e.~each $\omega \in \Omega$ represents a person.
Let $S \subset \Omega$ be the set of all \emph{smokers} and further 
\halfbline

\begin{minipage}{0.6\textwidth}
\begin{itemize}
\item $A_1 \subset \Omega$ be the persons of age $<10$
\item $A_2 \subset \Omega$ be the persons of $10 \leq$ age $<20$
\item $A_3 \subset \Omega$ be the persons of $20 \leq$ age $<50$
\item $A_4 \subset \Omega$ be the persons of $50 \leq $ age 
\end{itemize}
\end{minipage}~~~~~~~~~~%
\begin{minipage}{0.35\textwidth}
\includegraphics[width=0.99\textwidth]{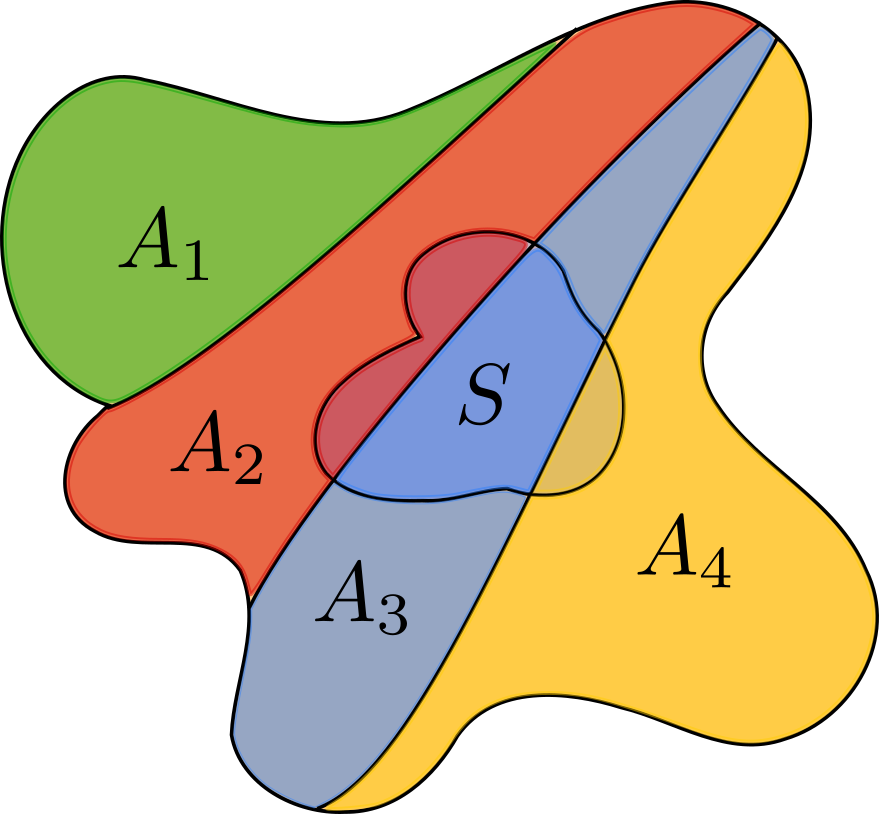}
\end{minipage}
\halfbline

Consequently, $\prob(S \cap A_1)$ is the probability that somebody smokes and is younger than $10$ years, 
$\prob(S \cap A_2)$ the probability that somebody smokes and is between $10$ and $20$, etc. 
Rather than keeping all these probabilities, we might decide to forget about age and compute the probability of smoking as 
$$
\prob(S) = \prob(S \cap A_1) + \prob(S \cap A_2) + \prob(S \cap A_3) + \prob(S \cap A_4)
$$
\end{example}
\end{tcolorbox}
\end{figure}

As a reasoning primitive, the law of total probability, or \emph{sum rule}, can be interpreted as \textbf{ignoring}, \textbf{accounting for unknown facts} or \textbf{forgetting}.
In particular, figure that we have elicited all the probabilities of the joint events $B \cap A_1, B \cap A_2, \dots$, that is, a rather detailed information status about whether $B$ and $A_1$ happen, $B$ and $A_2$ happen, etc.
By summing over these probabilities, we essentially forget (ignore, account for different possibilities) regarding the exclusive events $A_i$. 
Example \ref{ex:total_probability} illustrates the concept.

\subsection{Conditional Probability (The Product Rule)}
\label{sec:conditional_probability}

\begin{figure}[t]
\begin{tcolorbox}[width=\textwidth,]
\begin{example}[Conditional Probability Space]   
\label{ex:conditional_probability_space}
Let $\Omega$ be the sample space of a given probability space and $B$ an event we condition on.
The figure below illustrates how various events $A$ get translated into events $A' = A \cap B$ on the conditional probability space defined on $B$.
Conditional probability $\prob(A \cbar B) = {\prob(A \cap B) \over \prob(B)} = {\prob(A') \over \prob(B)}$ can now be interpreted as bona-fide probability measure on the sample space $B$.

\halfbline

\includegraphics[width=0.999\textwidth]{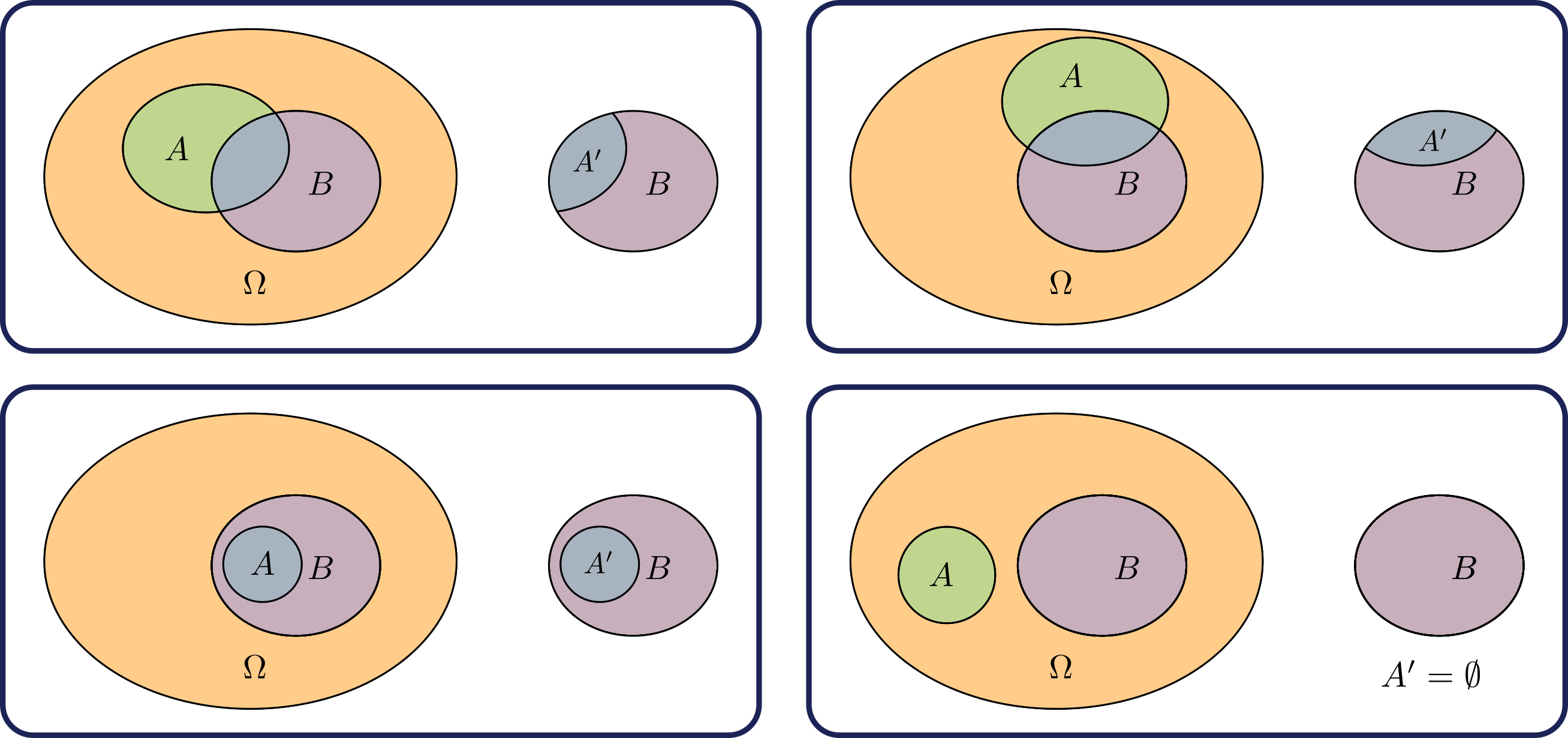}

\halfbline

One does not necessarily need to view $B$ as the new sample space, but one can also keep the entire $\Omega$.
Of course, the conditional probability for any set $A$ which does not overlap with $B$ must be $0$ then.
\end{example}
\end{tcolorbox}
\end{figure}

The second central reasoning primitive in probability is \emph{conditional probability}.
\begin{definition}[Conditional Probability]
Given an arbitrary probability space $(\Omega, \Sigma, \prob)$, let $A, B \in \Sigma$ be two arbitrary events, where $\prob(B) > 0$.
Then the \emph{conditional probability} of $A$ given $B$ is 
\begin{equation}
\label{eq:conditional_probability}
\prob(A \cbar B) = {\prob(A \cap B) \over \prob(B)}.
\end{equation}
\end{definition}
$\prob(A \cbar B)$ is simply the probability of $A$ after asserting that $\omega \in B$.
In particular, \mbox{$\prob(B \cbar B) = 1$}, as should be expected.
While the law of total probability (sum rule) shall be interpreted as ignoring, forgetting, or accounting for unknown facts, conditional probability shall be interpreted as \textbf{observing} or \textbf{injecting information}.
In particular, while the probability space $(\Omega, \Sigma, \prob)$ is our ``global'' mathematical model of uncertainty, conditional probability essentially ``rebases'' the probability space on $B$, and updates our model based on the information that $\omega \in B$. 
This is indeed correct as we can define a \emph{conditional probability space} as follows:
\begin{itemize}
\item Define $B$ as the new sample space.
\item Define a sigma-algebra $\Sigma'$ as
$$
\Sigma' = \{A \cap B \cbar A \in \Sigma\},
$$
i.e., the collection of the intersection of each original event with $B$.
It can be shown that $\Sigma'$ is indeed a sigma-algebra over $B$.
\item Define a (conditional) probability measure $\prob'$ on $\Sigma'$ as 
$$
\prob'(A') \coloneqq {\prob(A') \over \prob(B)} = {\prob(A \cap B) \over \prob(B)}
$$ 
Note that $\prob(A')$ is well-defined, since any $A'$ is given as the intersection $A' = A \cap B$ of events in the original $\Sigma$. 
Hence also $A' \in \Sigma$ and $\prob(A')$ is defined.
\end{itemize}
Then $(B, \Sigma', \prob')$ is again a probability space, conditional on $\omega \in B$.
Thus, we might see conditional probability as a tool that transforms our original probability space into a new one, accounting the provided information $\omega \in B$.
Example \ref{ex:conditional_probability_space} illustrates the concept.

\subsection{Bayes' Rule (The Rule of Inverse Probability)}
\label{sec:bayes_rule_measures}

The product rule immediately delivers \emph{Bayes' rule}.
Equation \eqref{eq:conditional_probability} can be written as
$$
{\prob(A \cap B)} = \prob(A \cbar B) \, \prob(B),
$$
and by symmetry we get 
$$
{\prob(A \cap B)} = \prob(B \cbar A) \, \prob(A),
$$
Equating the right hand sides and dividing by $\prob(B)$, we get Bayes' rule: 
\begin{equation}
\label{eq:Bayes1_events}
\prob(A \cbar B) = {\prob(B \cbar A) \, \prob(A) \over \prob(B)}.
\end{equation}
Bayes' rule is often called the \emph{law of inverse probability}, as \eqref{eq:Bayes1_events} essentially reverses the direction of reasoning.
Of course, by symmetry we also get 
\begin{equation}
\label{eq:Bayes2_events}
\prob(B \cbar A) = {\prob(A \cbar B) \, \prob(B) \over \prob(A)}.
\end{equation}

Bayes' rule is incredibly powerful, since one ``conditional direction'' (say $\prob(A \cbar B)$) is often easy to describe while one is actually interested in the other direction ($\prob(B \cbar A)$).
This is the general setting of \emph{inverse problems}, which are ubiquitous in science and engineering, such as in imaging, detection, and robotics.
Thus, Bayes' rule,  while being simple, describes the solution to the vast majority of these problems.
However, such \emph{conceptional} simplicity does not mean \emph{computational} simplicity, as most Bayesian problems are numerically challenging.

\subsection{(Conditional) Independence}  
\label{sec:conditional_independence}

Probabilities reflect the information content of an event.
Hence, conditional probabilities lead naturally to the notion of \emph{independence}.
When we consider an event $A$ whose probability does not change upon observing $B$, we define that $A$ and $B$ are independent of each other, denoted as $A \ci B$:
\begin{equation}   \label{eq:independence_events_1}
A \ci B  ~~~ \Leftrightarrow ~~~~ \prob(A \cbar B) = \prob(A)
\end{equation}
Due to the product rule $\prob(A \cap B) = \prob(A \cbar B) \, \prob(B)$ we also get
\begin{equation}   \label{eq:independence_events_2}
A \ci B  ~~~ \Leftrightarrow ~~~~ \prob(A \cap B) = \prob(A) \, \prob(B)
\end{equation}
and further, since ${\prob(A \cap B)} = \prob(B \cbar A) \, \prob(A)$,
\begin{equation}   \label{eq:independence_events_3}
A \ci B  ~~~ \Leftrightarrow ~~~~ \prob(B \cbar A) = \prob(B)
\end{equation}
which is the symmetrical case to \eqref{eq:independence_events_1}.
Hence, \eqref{eq:independence_events_1}, \eqref{eq:independence_events_2} and \eqref{eq:independence_events_3} are \emph{equivalent definitions} of independence.

Furthermore, independence can be conditional on some other event.
In particular, as illustrated Section~\ref{sec:conditional_probability}, conditional probability gives rise to a conditional probability space.
By simply applying the concept of independence to the conditional probability space, we get the notion of \emph{conditional independence}.
Specifically, let $A$, $B$ and $C$ be events, where $\prob(C) > 0$.
Then, $A$ and $B$ are \emph{conditionally independent}, conditionally on $C$, written as $A \ci B \cbar C$, if the following equivalent properties hold:
\begin{align}
\label{eq:ci_events1}
A \ci B \cbar C  ~~~ &\Leftrightarrow ~~~~ \prob(A \cbar B, C) = \prob(A \cbar C) \\
A \ci B \cbar C  ~~~ &\Leftrightarrow ~~~~ \prob(B \cbar A, C) = \prob(B \cbar C) \\
A \ci B \cbar C  ~~~ &\Leftrightarrow ~~~~ \prob(A, B \cbar C) = \prob(A \cbar C) \, \prob(B \cbar C) 
\end{align}
Note that they are the same as \eqref{eq:independence_events_1}, \eqref{eq:independence_events_2} and \eqref{eq:independence_events_3}, except that all probabilities are conditioned on $C$.

\section{Random Variables}
\label{sec:random_variables}

So far, we have discussed probability using its bare-bone measure theoretic definition:
We reflect our domain using a (possibly abstract) sample space $\Omega$, which is just a non-empty set.
The only thing which ``happens'' is that an atomic element $\omega$ is selected, and we don't know which one.
Our ignorance about which $\omega$ is selected is modelled by assigning real numbers---probabilities---between $0$ and $1$ to all kind of events in a consistent way, namely, that they add up for disjoint events.
Probability spaces are the ``Linux kernel'' of probability, a minimalist and simple theory which, however, unfolds into the rich structure of an intelligent calculus.
However, we usually don't want to work in the Linux kernel, but use an interpretable and high-level interface to work with.
The first step towards this interface are \emph{random variables} (RVs).

The basic idea of random variables is to map the abstract sample space $\Omega$ into something more interpretable, such as the real numbers $\reals$.
Thus, a random variable $X$ is really just a function $X \colon \Omega \mapsto \reals$, assigning each $\omega \in \Omega$ to some $X(\omega) \in \reals$.
Since the selection of $\omega$ is random, also the value returned by $X$ will be random, which is why a (deterministic) function defined on a probability space is called a ``random variable''.

Evidently, there is no reason to confine random variables to only return real numbers.
Rather, depending on the application, one would want them to return complex numbers, tuples of real numbers (vectors), graphs, functions, etc.
Thus, a random variable in general is just a transformation of one sample space $\Omega$ to some another (interpretable) space $\XS$.

In order to consistently work with probability represented by the probability space $(\Omega, \Sigma, \prob)$, we will require that $X$ is \emph{measurable}, which is defined as follows.
\begin{definition}[Measurable Function]
Let $(\Omega, \Sigma)$ and $(\XS, \Sigma_X)$ be measurable spaces and let $X \colon \Omega \mapsto \XS$ be a function.
$X$ is called \emph{measurable} if for each $B \in \Sigma_X$ it holds that 
$$
X^{-1}(B) \in \Sigma,
$$
where $X^{-1}$ is the \emph{preimage} of $B$ under $X$, i.e.\footnote{The symbol for pre-image is the same as for \emph{inverse function}, but these two are distinct concepts. In particular, a function might or might not have an inverse, but the pre-image is always defined.}
$$
X^{-1}(B) \coloneqq \{\omega \in \Omega \cbar X(\omega) \in B\}.
$$
\end{definition}
Measurability of a function requires that both domain $\Omega$ and co-domain $\XS$ are equipped with sigma-algebras $\Sigma$ and $\Sigma_X$, respectively, and that the pre-images of the events in $\Sigma_X$ are all contained in $\Sigma$.
It is thus a condition on the function itself as well as the measurable spaces $(\Omega, \Sigma)$ and $(\XS, \Sigma_X)$. 
Often one would fix the function of interest and design the sigma algebras such that the function indeed becomes measurable.

The defining part of a random variable $X$ is that it is a function defined on a probability space, mapping into a ``more interpretable space'' $\XS$, often called the \emph{state space} of $X$.
Since the selection of $\omega \in \Omega$ is uncertain also the value (state) of $X(\omega) \in \XS$ is uncertain.
This ``transformed'' uncertainty is described by the distribution of a random variable.

\subsection{Distribution of a Random Variable}

Given a probability space $(\Omega, \Sigma, \prob)$ and a random variable $X$ defined on this space and mapping into the measurable space $(\XS, \Sigma_X)$, we can define the following function:
$$
\prob_{X} \colon \Sigma_X \mapsto [0,1], ~~~
\text{where for any } B \in \Sigma_X \colon \prob_{X}(B) \coloneqq \prob(X^{-1}(B)).
$$
Function $\prob_X$ just assigns to each event $B \in \Sigma_X$ the probability which is assigned by $\prob$ to the preimage $X^{-1}(B)$, i.e.~the set of all $\omega$'s mapping to $B$.
Since the random variable is measurable it is guaranteed that $X^{-1}(B) \in \Sigma$, hence $\prob_X$ is properly defined.
It can be shown that $\prob_X$ is indeed a probability measure, now defined on $(\XS, \Sigma_X)$, and hence $(\XS,\Sigma_X,\prob_X)$ is a probability space.
$\prob_X$ is called the \emph{distribution induced by} $X$, the \emph{law of} $X$, or the \emph{pushforward measure} of $\prob$ under $X$, and it adequately describes our uncertainty about the value $X(\omega)$.

\subsection{Distribution Functions}
\label{sec:distribution_functions}

The introduction of random variables has just transformed the uncertainty over an ``uninterpretable'' space $\Omega$ into uncertainty over a ``more interpretable'' space $\XS$.
The latter is still described with a probability measure which is a rather cumbersome mathematical object. 
One also might ask why we did the hassle to first define a probability space $(\Omega,\Sigma,\prob)$ and a measurable function $X\colon \Omega \mapsto \XS$ just to end up with a new probability space $(\XS, \Sigma_X, \prob_X)$ describing the uncertain quantity $X$.
On this abstract level, it would be more straightforward to construct our model $(\XS, \Sigma_X, \prob_X)$ directly.
In the following, we will thus ignore the underlying probability space $(\Omega, \Sigma, \prob)$ and just aim to describe $(\XS, \Sigma_X, \prob_X)$ directly by the means of \emph{distribution functions}. 
The two most widely used distribution functions are \emph{probability mass functions} and \emph{probability density functions}.

\begin{definition}[Probability Mass Function (PMF)]
Let $X$ be a random variable with a countable state space $\XS$,\footnote{A random variable with countable state space is called \emph{discrete}.} equipped with sigma-algebra $\Sigma_X = 2^{\XS}$.
The \emph{probability mass function} of $X$ is defined as
$$
p_X \colon \XS \mapsto \reals, ~~~~~ p_X(x) = \prob_X(\{x\})
$$
which assigns to each state $x \in \XS$ the probability that $X = x$.
\end{definition}

A PMF is a simple and direct way to specify probability distributions.
In particular, the probability of any event $A \subseteq \XS$ is simply
\begin{equation}
\label{eq:prob_sumpmf}
\prob_{X}(A) = \sum_{x \in A} p_X(x),
\end{equation}
since any $A \subseteq \XS$ is countable and its probability must be given by the countable sum over the probabilities of the states in $A$, see Definition~\ref{def:probability_space}.
However, PMFs can only be defined for discrete random variables, as they assign probabilities in a ``discrete'' way to each atomic state $x \in \XS$.
This requires that we assume the power set $2^{\XS}$ as sigma-algebra, which is fine for discrete state spaces but does not work for uncountable state spaces, such as $\reals$ or $\mathbb{C}$.
For such random variables one commonly uses \emph{probability densities} instead.

\begin{definition}[Probability Density Function (PDF)]
Let $X$ be a random variable with the real numbers as state space, $\XS = \reals$, equipped with some sigma-algebra $\Sigma_X$ (usually the Borel sets) and probability measure $\prob_X$.
Let $p_X$ be a function mapping from $\XS$ to the non-negative real numbers, $p_X \colon \XS \mapsto [0, \infty]$ with the property that 
\begin{equation}
\label{eq:probability_density}
\forall A \in \Sigma_X \colon 
\prob_X(A) = \int_A p_X(x) \, \mathrm{d}x, 
\end{equation}
that is, for any event $A \in \Sigma_X$ the probability $\prob_X(A)$ is given by the integral of $p_X$ over $A$.
Then $p_X$ is called a \emph{probability density function} of $X$, also just \emph{probability density} or \emph{density}.
A random variable which admits a density is called \emph{continuous}.
\end{definition}

The probability density, if it exists, is not unique.
In particular, any density $p_X$ can be changed at a countable collection of points without changing the integral \eqref{eq:probability_density} for any $A$.

Note that both PMFs and PDFs are denoted by the same symbol $p_X$.
The intuitive reason for this is that both serve the same purpose, namely to represent a probability measure $\prob_X$ of a random variable $X$, either via a sum \eqref{eq:prob_sumpmf} or an integral  \eqref{eq:probability_density}.
Moreover, within a measure-theoretic treatment of probability these two concepts become in fact the same.
Specifically, when $\mu$ is any measure defined on $(\XS, \Sigma_X)$ and $p_X \colon \XS \mapsto [0, \infty]$ is a measurable function with 
\begin{equation}
\label{eq:radon_nikodym}
\forall A \in \Sigma_X \colon 
\prob_X(A) = \int_A p_X \, \mathrm{d} \mu
\end{equation}   
then $p_X$ is called a \emph{Radon–Nikodym derivative} of $\prob_X$ with respect to $\mu$.
In that sense, a PMF is just a Radon-Nikodym derivative with respect to the \emph{counting measure} and a PDF is just a Radon-Nikodym derivative with respect to Lebesgue measure (capturing the notion of length, area, volume, etc.). 
For brevity, we might just use the shorter term \emph{density} for $p_X$ and also be lose about whether we speak about a discrete random variable (hence using sums) or a continuous random variable (hence using integrals).

\subsection{Multivariate Random Variables and Joint Distributions}
\label{sec:joint_distribution_functions}

So far we were concerned mainly with univariate random variables, returning a ``single'' value.
However, the usual tasks in science and engineering are to describe how \emph{multiple} quantities are entangled with each other. 

We again consider some (abstract) probability space $(\Omega, \Sigma, \prob)$ and define now \emph{multiple} random variables on it.
We might call them $X$, $Y$, $Z$ or more systematically enumerate them as $X_1, X_2, \dots, X_D$, where $D \in \naturals$, 
and let $\XS_1, \XS_2, \dots, \XS_D$ be their state spaces.
Now, since the mental picture is that some atomic event $\omega \in \Omega$ is selected, where $\prob$ describes our ignorance about this selection, the returned values $X_1(\omega), X_2(\omega), \dots, X_D(\omega)$ will be uncertain but (in general) also correlated with each other.

The discussion of multivariate random variables really does not add much mathematical depth to univariate random variables. 
In fact, one simply defines a \emph{random tuple} $\X$ (also often referred to as \emph{random vector} or \emph{set of random variables}) as the function 
$$
\X \colon \Omega \mapsto \XXS, ~~~~~
\X(\omega) \coloneqq (X_1(\omega), X_2(\omega), \dots, X_D(\omega))
$$ 
mapping from $\Omega$ to the Cartesian product of the state spaces $\XXS \coloneqq \bigtimes_{i=1}^D \XS_i$ and whose $i\textsuperscript{th}$ component is the univariate random variable $X_i$.
In order to assert that $\X$ is a (vector-valued) random variable one needs to ensure that $\X$ is measurable whenever the univariate random variables $X_1, X_2, \dots, X_D$ are measurable.
Indeed, one can show that $\X$ is measurable with respect to the so-called \emph{product sigma-algebra} $\Sigma_{\X}$ which is the sigma-algebra generated by $E = \left \{\bigtimes_{i=1}^D A_i \cbar A_i \in \XS_i \right \}$, i.e.~the set of all Cartesian products of any events of the univariate random variables.
Thus, $\X$ can be understood as a standard random variable, albeit one mapping to the multi-dimensional space $\XXS$ and equipped with product sigma-algebra $\Sigma_{\X}$.

The concept of distribution functions carries over to multivariate random variables as follows.
\begin{definition}[Joint Probability Mass Function (Joint PMF)]
Let $\X = (X_1, \dots, X_D)$ be a random vector, consisting of discrete univariate random variables $X_1, \dots, X_D$ with countable state spaces $\XS_1, \dots, \XS_D$.
The \emph{joint probability mass function} is defined as
$$
p_{\X} \colon \XXS \mapsto [0,1], ~~~~~ p_{\X}(\x) = \prob_{\X}(\{\x\})
$$
which assigns to each joint state $\x$ the probability that $\X = \x$.
\end{definition}
Similarly to univariate random variables the joint PMF describes the probability measure of $\X$ via
\begin{equation}
\label{eq:multivariate_pmf_sum}
\prob_{\X}(A) = \sum_{\x \in A} p_{\X}(\x). 
\end{equation}
for any event $A \in \Sigma_{\X}$.
Furthermore, also the concept of PDF carries over to the multivariate case.
\begin{definition}[Joint Probability Density Function (Joint PDF)]
Let $\X = (X_1, \dots, X_D)$ be a random vector, consisting of univariate random variables $X_1, \dots, X_D$ each with state space $\reals$ and equipped with the Borel sigma-algebra.
Let $p_{\X}$ be a function mapping from $ \reals^D$ to the non-negative real numbers, $p_{\X} \colon \reals^D \mapsto [0, \infty]$, with the property that 
\begin{equation}
\label{eq:multi_probability_density}
\prob_{\X}(A) = \int_A p_{\X}(\x) \, \mathrm{d}\x, 
\end{equation}
for any event (Borel set) $A \subseteq \reals^D$.
Then $p_{\X}$ is called a \emph{joint} probability density of $\X$.
\end{definition}

The integral \eqref{eq:multi_probability_density} (or sum \eqref{eq:multivariate_pmf_sum}) is the defining part of a joint distribution function $p_{\X}$: when integrating (or summing) it over an event $A \in \Sigma_{\X}$, it needs to return $\prob_{\X}(A)$.
From this perspective, one can also define \emph{hybrid} distribution functions over sets of random variables $\X$ which contains both discrete and continuous ones.
\begin{definition}[Hybrid Distribution Function]
Let $\X = (X_1, \dots, X_D)$ be a random vector where $\X' = (X_1, \dots, X_K)$ are discrete random variables and $\X'' = (X_{K+1}, \dots, X_D)$ are continuous random variables,\footnote{That the first $K$ random variables are discrete and the others continuous is merely an assumption for convenience of notation, which can be made without loss of generality.} and let $\XXS = \bigtimes_{i=1}^D \XS_i$ be their joint state space.

For any event $A \in \Sigma_{\X}$ let 
$$
A' \coloneqq \{ (x_1, \dots, x_K) \cbar (x_1, \dots, x_D) \in A\} 
$$
i.e., the set of all states assumed by the discrete random variables in $A$.
Furthermore, for each $\x' \in A'$ let
$$
A''_{\x'} \coloneqq \{ (x_{K+1}, \dots, x_D) \cbar (x_1, \dots, x_D) \in A \land (x_1, \dots, x_K) = \x' \}
$$
i.e., the set of all states assumed by the continuous random variables assumed in $A$, whenever the discrete random variables assume state $\x'$.

Let $p_{\X} \colon \XXS \mapsto [0, \infty]$ be a function with the property that for any $A \in \Sigma_{\X}$
\begin{equation}
\label{eq:hybrid_distribution}
\prob_{\X}(A) = \sum_{\x' \in A'} \int_{A''_{\x'}} p_{\X}(\x', \x'') \, \mathrm{d}\x''.
\end{equation}
Then $p_{\X}$ is called a \emph{hybrid distribution function} over $\X$.
\end{definition}

The definition of hybrid distribution functions is a bit cumbersome, as it requires us to split random variables and their state spaces into discrete and continuous ones.
Furthermore, all the joint distributions---joint PMFs, joint PDFs, and hybrid distribution functions---can be understood as densities in a measure-theoretic sense, as in \eqref{eq:radon_nikodym}.
The only difference between these is just the base measure: for discrete random variables one uses a counting measure, for continuous random variables one typically uses the Lebesgue measure, and for the hybrid case one would use an adequate product measure of counting and Lebesgue measure.

Hence, similarly as in the univariate case we use the same symbol $p_{\X}$ for joint PMFs, joint PDFs and hybrid distribution functions, as they all serve the same purpose---describing a probability measure $\prob_{\X}$ by integrating (or summing) $p_{\X}$ over some event $A$.

%% file: Chapter3/chapter3.tex
\chapter{Probabilistic Inference}   
\label{chap:probabilistic_reasoning}

As discussed in Section~\ref{sec:what_is_ai}, a core property of any AI system is to have some representation of a \emph{knowledge base}.
This role is now fulfilled by distribution functions $\p_{\X}$ introduced in the last chapter, representing both  \emph{dependencies between random variables} as well as our \emph{uncertainty} about them.
Besides representing our knowledge as a distribution function $p_{\X}$ we ultimately are interested in \emph{reasoning}, or \emph{inference}, i.e.~answering queries of interest.
Probability naturally offers \emph{inference routines} in order to perform reasoning under uncertainty. 
The two most important routines are \textbf{marginalization (the sum rule)} and \textbf{conditioning (the product rule)}, where the latter just uses the former as a sub-routine.
Hence, marginalization is really the core business of probabilistic inference.

Besides these two central inference routines there are several other fundamental operations one would want to do with probability distributions, in particular taking \textbf{expectations} and performing \textbf{optimization}, i.e.~taking maxima or minima of functions of interest.
As it turns out, however, most models mentioned above struggle with computing these queries, as they are typically NP-hard problems in these. 
\textbf{Probabilistic circuits (PCs)}, which we introduce in Chapter \ref{chap:probabilistic_circuits}, will allow to compute these queries \textbf{exactly} and \textbf{efficiently}.

Before diving into PCs, however, in this chapter we shall study the elegant reasoning calculus probability has to offer, disregarding computational issues.
In fact, one of the things which make probabilistic models so interesting is the tension between the \textbf{conceptional} simplicity of the calculus---it is in principle clear \emph{what} shall be done---and the \textbf{computational hardness} involved when implementing the calculus.

\section{Marginalization (The Sum Rule)}

Let a distribution function $p_{\X}$ for a random vector $\X = (X_1,\dots,X_D)$ be given and let $\Y = (Y_1,\dots,Y_{D'})$ and $\Z=(Z_1,\dots,Z_{D''})$ be a partition of $\X$, i.e.~$\Y \cup \Z = \X$ and $\X \cap \Z = \emptyset$.
We assume without loss of generality that $\Y$ are the first $D'$ entries in $\X$ and $\Z$ the remaining $D'' = D-D'$ entries, i.e.~
$$
\X = (Y_1, \dots, Y_{D'}, Z_{1}, \dots, Z_{D''}).
$$
The \emph{marginal distribution} $p_{\Y}$ is given as 
\begin{equation}
\label{eq:marginal_manyintegrals}
p_{\Y}(y_1, \dots, y_{D'}) = \int \dots \int p_{\X}(y_1, \dots, y_{D'}, z_1, \dots, z_{D''}) \, \mathrm{d}z_1 \dots \mathrm{d}z_{D''}, 
\end{equation}
that is, we \emph{integrate out} or \emph{marginalize}\footnote{Recall that we do not make much difference whether random variables are discrete or continuous and that for discrete random variables we simply replace an integrals in \eqref{eq:marginal_manyintegrals} by sums.} all random variables in $\Z$, in order to get a ``reduced model'' $p_{\Y}$ only over random variables $\Y$.
We often write \eqref{eq:marginal_manyintegrals} more compactly as 
\begin{equation}
\label{eq:marginal}
p_{\Y}(\y) = \int p_{\X}(\y, \z) \, \mathrm{d}\z, 
\end{equation}
where we ``unpack'' the values $\y, \z$ as arguments in $p_{\X}(y_1, \dots, y_{D'}, z_1, \dots, z_{D''})$ and $\y$ as arguments in $p_{\Y}(y_1, \dots, y_{D'})$.
While the full joint distribution $p_{\X}$ captures the dependency structure and uncertainties among \emph{all} $\X$, the marginal \eqref{eq:marginal} accurately describes dependencies and uncertainties among subset $\Y$, while having eliminated $\Z$ from consideration.
Thus, the marginalization operation \eqref{eq:marginal} shall be interpreted as the reasoning primitive \textbf{ignoring} or \textbf{accounting for unknown quantities} $\Z$. 
Since this happens by integrating or summing over all possible states of the unknown quantity $\Z$, this operation is also often called the \textbf{sum rule}.
Basically, it is the \emph{law of total probability} (see Section \ref{sec:law_total_probability}) applied to distribution functions.

A few things are worth mentioning.
First, note that $p_{\Y}$ is in general a \emph{joint} distribution, but one over a subset of $\X$.
Of course, whenever $\Y = \{Y_i\}$ is singleton, $p_{\Y}$ becomes a univariate distribution $p_{Y_i}$.  
The marginal distribution is indeed ``the'' distribution over $\Y$ and not some kind of approximation.
Specifically, just as $p_{\X}$ accurately describes a set of random variables $\X$, $p_{\Y}$ accurately describes $\Y$, ``as if $\Z$ had never been considered''.

Also the extreme case of marginalization where all variables $\X$ are marginalized is interesting, as it delivers the \textit{normalization constant} or \textit{partition function}
\begin{equation}
	\mathcal{Z} = \int p_{\X}(\x) \, \mathrm{d}\x.
\end{equation}
The normalization constant is important for \textit{unnormalized models} which represent distribution functions via some nonnegative function $f_{\X}$:
\begin{equation}
	p_{\X}(\x) \coloneqq {f_{\X}(\x) \over \mathcal{Z}}, ~~~~~ \mathcal{Z} = \int f_{\X}(\x) \, \mathrm{d}\x.
\end{equation}

Furthermore note that marginalization is a \textit{consistent} reasoning pattern, meaning that one always arrives at the same result for any sequence of marginalization operations expressing the same marginal.
For example, when starting with a joint $p_{\X,\Y,\Z}$ we will arrive at the same marginal $p_{\X}$ whether we are (i) first marginalizing $\Z$ (yielding $p_{\X,\Y}$) and then $\Y$, or (ii) first marginalizing $\Y$ (yielding $p_{\X,\Z}$) and then $\Z$.

\section{Conditioning (The Product Rule)}

Let again a distribution function $p_{\X}$ for a set of random variables $\X = (X_1,\dots,X_D)$ be given and let $\Y = (Y_1,\dots,Y_{D'})$ and $\Z=(Z_1,\dots,Z_{D''})$ be a partition of $\X$.
Given a joint state $\z$ for $\Z$, the \emph{conditional distribution} of $\Y$ given $\z$ is 
\begin{equation}
\label{eq:conditional}
p_{\Y|\Z}(\y \cbar \z) 
= {p_{\X}(\y, \z) \over p_{\Z}(\z)}
= {p_{\X}(\y, \z) \over {\int p_{\X}(\y, \z) \, \mathrm{d}\y}},
\end{equation} 
i.e., the ratio of the joint distribution $p_{\X}$ and the marginal $p_{\Z}$, where the latter is derived from $p_{\X}$ using the sum rule.
The conditional distribution is the concept of conditional probability (see Section \ref{sec:conditional_probability}) applied to distribution functions.

A natural way to interpret $p_{\Y|\Z}$ is as a \emph{family of distribution functions} over $\Y$, \emph{indexed} by the states $\z \in \mathbfcal{Z}$.  
While the full joint distribution $p_{\X}$ captures the dependencies and uncertainties among \emph{all} random variables $\X$, the conditional distribution $p_{\Y|\Z}$ describes dependencies and uncertainties among $\Y$ upon establishing the event $\Z = \z$.
Hence, the conditioning operation \eqref{eq:conditional} shall be interpreted as the reasoning primitive \textbf{observing} or \textbf{injecting evidence} (that $\Z = \z$) and updating the distribution over the remaining random variables $\Y$.

Equation \eqref{eq:conditional} can also be written as 
\begin{equation}
	\label{eq:product_rule}
	{p_{\X}(\y, \z)} = 
	p_{\Y|\Z}(\y \cbar \z) \, p_{\Z}(\z), 
\end{equation}  
telling us that the joint distribution over $\Y$ and $\Z$ is given as the product of the marginal distribution over $\Z$ and the conditional distribution over $\Y$ given $\Z$.
This essentially represents a ``reasoning step-by-step'' pattern: in order to reason about \emph{both} $\Y$ and $\Z$, first reason about \emph{only} $\Z$ and subsequently about $\Y$ \emph{given} $\Z$. 
Since the joint distribution is given as a product of marginal and conditional distributions, Equation \eqref{eq:product_rule} is also referred to as the \textbf{product rule}.

There are some technical subtleties with conditioning, as in order for \eqref{eq:conditional} to be well-defined, the denominator $p_{\Z}(\z)$ needs to be non-zero.
When $\Z$ is discrete, however, we might just assume an arbitrary conditional distribution $p_{\Y|\Z}(\cdot \cbar \z)$ whenever $p_{\Z}(\z)=0$.
To see this note that for any $A \subseteq \YYS$ we have
$$
\int_{A} p_{\Y,\Z}(\y, \z) \, \mathrm{d}\y 
\leq 
\int_{\YYS} p_{\Y,\Z}(\y, \z) \, \mathrm{d}\y 
= 
p_{\Z}(\z) = 0.
$$
Thus, for any arbitrary $p_{\Y|\Z}(\cdot \cbar \z)$ we get 
$$
\int_{A} p_{\Y,\Z}(\y, \z) \, \mathrm{d}\y
= 
\int_{A} p_{\Y|\Z}(\y \cbar \z) \, p(\z) \, \mathrm{d}\y 
= 
p(\z) \, \overbrace{\int_{A} p_{\Y|\Z}(\y \cbar \z) \, \mathrm{d}\y }^{\leq 1}
= 0,
$$
as required.

When $\Z$ is continuous, and hence $p_{\Z}$ is a density, Equation \eqref{eq:conditional} yields a conditional distribution function which is conditional on an event of probability $0$.
Yet, conditioning is well-defined when the marginal density $p_{\Z}(\z)$ is non-zero and continuous at $\z$.

Conditioning is, like marginalization, a consistent reasoning operation, meaning that the order of conditioning operations does not matter.
For example, when we start with a joint $p_{\X,\Y,\Z}$ and want to compute $p_{\X|\Y,\Z}$, we might first condition on $\Y$, yielding $p_{\X,\Z|\Y}$, and subsequently on $\Z$, or, we we condition first on $\Z$ and subsequently on $\Y$.
To see this, note that the conditional 
$$
p_{\X,\Z|\Y}(\x,\z\cbar \y)  = {p_{\X,\Y,\Z}(\x,\y,\z) \over p_{\Y}(\y)},
$$
specifies a family of joint distribution over $\X$ and $\Z$ for every value $\y$.
Conditioning each of these conditional distributions additionally on $\Z$ delivers 
$$
{p_{\X,\Z|\Y}(\x,\z\cbar \y) \over p_{\Z|\Y}(\z\cbar \y) }
 = 
{p_{\X,\Y,\Z}(\x,\y,\z) \over p_{\Y}(\y)}
\overbrace{{p_{\Y}(\y) \over \int p_{\X,\Y,\Z}(\x,\y,\z) \, \mathrm{d}\x}}^{1 /p_{\Z|\Y}(\z\cbar \y)}
=  
{p_{\X,\Y,\Z}(\x,\y,\z) \over p_{\Y,\Z}(\y,\z)}
=
p_{\X|\Y,\Z}(\x \cbar \y,\z)
$$
Similarly, by swapping the role of $\Y$ and $\Z$, we will get the same result by first condition on $\Z$ and then on $\Y$.

Conditioning and marginalization are also compatible with each other.
For example, when we start with a joint $p_{\X,\Y,\Z}$ and want to compute $p_{\X|\Z}$, we might 
\begin{itemize}
\item first condition on $\Z$ (yielding $p_{\X,\Y|\Z}$) and then marginalize $\Y$, or
\item  we might first marginalize $\Y$ (yielding $p_{\X,\Z}$) and then condition on $\Z$.
\end{itemize}
In general, any two sequences of marginalization and conditioning operations leading to the same marginal and/or conditional will deliver consistent results.\footnote{This insight is referring to the exact mathematical definition of marginals and conditionals. When performing \emph{approximate} inference the results might of course depend strongly on the order of operations due to numerical and other reasons. In fact, a lot of interesting methods in probabilistic inference is based on clever choices of picking such order and beneficial computational techniques.}

Hence, with the sum and product rule we have established the central ``inference machinery'' of probabilistic reasoning: we start with a joint distribution $p_{\X}$ describing our domain of discourse. 
Then we select \emph{variables we want to infer} $\Y \subseteq \X$, depict \emph{variables we know} $\Z \subseteq \X$, and \emph{ignore the rest} $\X \setminus (\Y \cup \Z)$.
It is worth to embrace the simplicity and elegance of this calculus, allowing us to (repeatedly) apply just two rules to transform our knowledge base $p_{\X}$ into the object of desire, $p_{\Y|\Z}$.

Before we proceed let us simplify a bit our notation. 
In particular, we have now introduced symbols like $p_{\X}$ and $p_{\Y,\Z}$ for joint distributions and symbols like $p_{\Z|\X}$ and $p_{\Y,\X|\Z}$ for conditional distributions.
Distinguishing these different marginals and conditionals is essential for our intelligent calculus, but using these subscripts quickly becomes cluttered when, for example, describing the product rule:
$$
p_{\X}(\y, \z) = p_{\Y|\Z}(\y \cbar \z) \, p_{\Z}(\z).
$$
These subscripts are redundant, since it is clear from the arguments to which random variables the distribution belongs, and also whether it is a conditional or an unconditional distribution.
We might just write more compactly
$$
p(\y, \z) = p(\y \cbar \z) \, p(\z).
$$
Hence, we will from now omit subscripts of distribution functions when their nature is clear form the arguments.

\section{The Chain Rule}

As mentioned above, the product rule essentially implements a \emph{two-step} reasoning pattern.
By applying the product rule repeatedly, we arrive at the \emph{chain rule}, which implements a \emph{multi-step} reasoning pattern.  
Let $\X$ be a set of random variables which is partitioned into sets $\X_1, \dots, \X_K$. 
In the following, note that it does not matter whether any $\X_i$, $i \in \{1,\dots,K\}$, contains only a single or several random variables, i.e.~the chain rule applies regardless whether we consider single random variables or ``blocks'' of random variables.   
Using the product rule \emph{repeatedly}, we can now write $p_{\X}$ as 
\begin{align}
\nonumber
p(\x) & = p(\x_1, \x_2,\x_3,\x_4, \dots,\x_{K})  \\
\nonumber
& = p(\x_2,\x_3,\x_4, \dots,\x_{K} \cbar \x_1) \, p(\x_1) \\
\nonumber
& = p(\x_3,\x_4, \dots,\x_{K} \cbar \x_1, \x_2) \, p(\x_2 \cbar \x_1)\, p(\x_1) \\
\nonumber
&=  p(\x_4, \dots,\x_{K} \cbar \x_1, \x_2, \x_3) \, p(\x_3 \cbar \x_1,\x_2) \, p(\x_2 \cbar \x_1)\, p(\x_1) \\
\nonumber
& ~~~~~ \dots \\
\nonumber
& = p(\x_{K} \cbar \x_1, \dots, \x_{K-1}) p(\x_{K-1} \cbar \x_1, \dots, \x_{K-2}) \, \dots \, p(\x_3 \cbar \x_1,\x_2) \, 
p(\x_2 \cbar \x_1)\, p(\x_1) \\
& = \prod_{k=1}^{K} p(\x_k \cbar \x_1, \dots, \x_{k-1}).
\label{eq:chain_rule}
\end{align}
In \eqref{eq:chain_rule} the factor $p(\x_k \cbar \x_1, \dots, \x_{k-1})$ for $k=1$ shall be interpreted as $p(\x_1)$.
This factorization of the joint $p_{\X}$ is known as the \emph{chain rule} and represents ``step-by-step reasoning'' over multiple steps, as it tells us that the joint distribution over $\X_1, \dots, \X_K$ is constructed as the marginal over $\X_1$ times the conditional over $\X_2$ given $\X_1$ times the conditional $\X_3$ given $\X_1,\X_2$, etc.

We are not confined to use the chain rule using the ordering $1, 2, \dots, K$, but are free to use any ordering we like.
Specifically, for any permutation $i_1, \dots, i_K$ of the integers $\{1,\dots,K\}$, we can write $p_{\X}$ as 
\begin{align}
p(\x) = \prod_{k=1}^{K} p(\x_{i_k} \cbar \x_{i_1}, \dots, \x_{i_k-1}).
\label{eq:chain_rule_ordering}
\end{align}
For example, we might factorize a joint distribution over $\X_1,\X_2,\X_3$ in various ways:
\begin{align*}
p(\x_1, \x_2, \x_3) 
& = p(\x_3 \cbar \x_2,\x_1) \, p(\x_1 \cbar \x_2) \, p(\x_2)  \\
& = p(\x_2 \cbar \x_1,\x_3) \, p(\x_3 \cbar \x_1) \, p(\x_1)  \\
& = p(\x_1 \cbar \x_3,\x_2) \, p(\x_2 \cbar \x_3) \, p(\x_3)  \\
& ~~~~~ \text{etc.} 
\end{align*}
Thus, rather then a single chain rule, for a given partitioning $\X_1, \dots, \X_
K$ there really exist $K!$ different chain rules.
The chain rule allows to factor and analyse joint distributions in a flexible way and can be considered as a ``Swiss army knife'' of probabilistic inference.

Conversely, reading the chain rule in the other direction, we can \emph{construct} expressive joint distributions by providing a marginal or conditional for each $\X_k$ and multiplying these together.
Here, the conditionals need to obey \emph{some} ordering of the $\X_k$'s, meaning that an $\X_k$ can only appear as conditioned in some conditional distribution $p_{\X_l|\dots, \X_k, \dots}$ when $\X_k$ comes before $\X_l$ in this order.
As long as there exists such an order, the chain rule leads to a proper joint distribution $p_{\X}$.
This is the main working principle both behind classical Bayesian networks \cite{koller2009probabilistic} and autoregressive distribution estimators \cite{larochelle2011neural}.

\section{Bayes Rule (Law of Inverse Probability)}
\label{sec:bayes_rule}

Another consequence of the product rule is \emph{Bayes rule} or the \emph{rule of inverse probability}, which has been discussed already in Section \ref{sec:bayes_rule_measures} on the level of probability measures.
When considering two (sets of) random variables $\X,\Y$, we might write the joint as 
$$
p(\x,\y) = p(\y \cbar \x) \, p(\x)
$$
or also 
$$
p(\x,\y) = p(\x \cbar \y) \, p(\y).
$$
By equating the right hand sides and dividing by $p(\x)$ we yield Bayes rule
\begin{equation}
\label{eq:bayes_rule}
 p(\y \cbar \x) = {p(\x \cbar \y) \, p(\y) \over p(\x)} = {p(\x \cbar \y) \, p(\y) \over \int p(\x \cbar \y) \, p(\y) \, \mathrm{d}\y},
\end{equation}
which allows us to change the ``direction of reasoning'' from $\x \leftarrow \y$  to $\y \leftarrow \x$.
Bayes rule is in particular prominent in applications were $\Y$ is some unobserved (latent) quantity, such as an unknown model parameter, the truth status (\emph{true}, \emph{false}) of a hypothesis, or some other (assumed) latent structure underlying an application, while $\X$ is an observed quantity (data).
In such settings, it is often natural to describe how the data $\X$ emerges from the latent quantity $\Y$ via some conditional $p_{\X|\Y}$, usually called the \emph{likelihood}.
Together with a marginal distribution $p_{\Y}$, usually called the \emph{prior}, Bayes rule \eqref{eq:bayes_rule} allows us to infer the conditional $p_{\Y|\X}$ of interest, called the \emph{posterior}.
In this way, Bayes rule captures a wide range of inverse problems in science and engineering.

\section{(Conditional) Independence}
\label{sec:conditional_independence_distributions}

The concept of independence discussed in Section~\ref{sec:conditional_independence} naturally carries over to distributions functions.
The conditional distribution $p_{\Y | \X}$ represents our information about $\Y$ when we have learned the value for $\X$.
If this this is the same as when we do \emph{not} know the value of $\X$, i.e.~the marginal $p_{\Y}$, we conclude that $\Y$ and $\X$ are independent. 
Consequently, we define $\X \ci \Y$, when the following equivalent conditions hold for all values $\x,\y$:
\begin{itemize}
\item $p(\x \cbar \y) = p(\x)$
\item $p(\y \cbar \x) = p(\y)$
\item $p(\x, \y) = p(\x) \, p(\y)$
\end{itemize}

Independence can occur or disappear in the context of another set of random variables $\Z$.
In particular, we say that $\X$ and $\Y$ are \emph{conditionally independent} given $\Z$, written $\X \ci \Y \cbar \Z$, if the following equivalent conditions hold for all values $\x,\y,\z$:
\begin{itemize}
\item $p(\x \cbar \y, \z) = p(\x \cbar \z)$
\item $p(\y \cbar \x, \z) = p(\y \cbar \z)$
\item $p(\x, \y \cbar \z) = p(\x \cbar \z) \, p(\y \cbar \z)$
\end{itemize}

\section{Expectations}

So far, we have established that probability allows us to \emph{represent knowledge in the form of a joint distribution} and provides  two simple inference rules to process our knowledge, \emph{the sum and the product rule}.
We have also established the chain rule and Bayes rule, which, however, are really just variants of the product rule.
All this calculus does is processing distributions into other distributions, as illustrated in the following example.

\begin{tcolorbox}[width=\textwidth,]
\begin{example}
\label{ex:blood_flow}
Assume we have access to the joint distribution over $21$ random variables $Y, X_1, \dots, X_{20}$, where $Y$ is some medical parameter of interest (e.g.~cerebral blood flow), and $X_1, \dots, X_{20}$ are some measurements related to $Y$ (e.g.~features derived from $20$ EEG channels).
Further, assume that for a particular patient we have measured (observed) $X_1 = x_1, X_2=x_2,\dots$, but $X_{17}$ and $X_{19}$ have not been measured (because for these two the corresponding EEG sensors have dropped out).
Hence, employing the sum and product rule we compute 
\begin{equation}
	\label{eq:infer_bloodflow}
	p(y \cbar x_1,\dots,x_{16},x_{18},x_{20}) = 
	{
		\int \int p(y, x_1,\dots,x_{16},x_{17},x_{18},x_{19},x_{20}) \,\mathrm{d}{x_{17}}\,\mathrm{d}{x_{19}}
		\over 
		\int \int \int p(y, x_1,\dots,x_{16},x_{17},x_{18},x_{19},x_{20}) \,\mathrm{d}{x_{17}}\,\mathrm{d}{x_{19}} \, \mathrm{d}y
	},
\end{equation}
which is our inferred distribution about $Y$, incorporating all available information.
\end{example}
\end{tcolorbox}

In Example~\ref{ex:blood_flow}, one might wonder what shall be done next with $p_{Y|X_1,\dots,X_{16},X_{18},X_{20}}$, a potentially very complicated and multimodal distribution?\footnote{The \emph{mode} of a distribution is the value of $y$ for which $p(y)$ is maximal. A ``bumpy'' distribution function which has more than one (local) maxima is called \emph{multimodal}. Intuitively, multimodality signals ambiguity about $Y$.}
How do we get a \emph{concrete} prediction for $Y$?
The first response to these concerns is that \emph{nothing} should be done, since the conditional distribution over $Y$ \emph{is} our concrete prediction.
More precisely, \emph{if} (i) the joint $p_{Y,X_1,\dots,X_{20}}$ is indeed the true distribution over all involved quantities, or at least a sufficiently good approximation of it, (ii) the observed values $x_1, \dots, x_{16}, x_{18}, x_{20}$ are indeed all the information we have\footnote{Note that here we assume that $X_{17}$ and $X_{19}$ are \emph{missing at random}, meaning that the fact that they are missing is independent of $Y$ given the observed RVs
\cite{rubin1976inference}.
In this case \eqref{eq:infer_bloodflow} is an exact conditional of $Y$ given the available data.
If they were not missing at random, one would need to account for this fact, for example by explicitly incorporating the missingness process in the model.} and (iii) \eqref{eq:infer_bloodflow} is solved exactly, there is nothing more that we can do.
The conditional $p_{Y|X_1,\dots,X_{16},X_{18},X_{20}}$ captures everything we can know about $Y$ based on the available information, or actually, it captures what we \emph{don't} know, namely our uncertainty about $Y$.
If the distribution is unimodal and very concentrated we are pretty confident about the value of $Y$, while if it is spread out and multimodal we are pretty uncertain.
Keeping the whole distribution of $Y$ means that \emph{we know what we don't know}.

However, humans tend to be uncomfortable with this situation.
They want to call the shots and know a concrete single value for $Y$. 
Often, committing to a single value is also an external requirement, forcing us to make a \emph{decision}.
The basic tool to do so is \textbf{expectation}, which is simply the average value a random variable assumes:
\begin{equation}
\label{eq:expectation_1}
\E[X] \coloneqq \int p(x) x \, \mathrm{d}x,
\end{equation}
where the integral is taken over the state space of $X$.
For discrete random variables, the integral is replaced by a sum.
For random vectors the expectation is a vector-valued quantity:
\begin{equation}
\label{eq:expectation}
\E[\X] \coloneqq \int p(\x) \x \, \mathrm{d}\x.
\end{equation}

Expectations can be naturally applied to conditional distributions.
When $p_{\Y|\Z}$ is a conditional distribution, the conditional expectation of $\Y$ given the event $\Z=\z$ is 
\begin{equation}
	\E[\Y\cbar \z] = \int p(\y \cbar \z) \,\y \, \mathrm{d}\y. 
\end{equation} 
In Example~\ref{ex:blood_flow}, computing $\E[Y \cbar x_1,\dots,x_{16}, x_{18},x_{20}]$ is one particular way to decide a value for $Y$.
However, being an average does not necessarily make it a typical value of the distribution, as illustrated in Example~\ref{ex:expectation_of_bernoulli}.

\begin{figure}
\begin{tcolorbox}[width=\textwidth,]
\begin{example}[Expectation of a Bernoulli]
\label{ex:expectation_of_bernoulli}
Consider a \emph{Bernoulli} distribution over a random variable $X$ with state space $\XS = \{0,1\}$, assigning probabilities $p_X(1) = \theta$ and $p_X(0) = 1-\theta$, where $\theta$ is a single parameter called the \emph{success probability}.
When $\theta=0.5$, the expectation is $\E[X] = 0.5$ which is a value never assumed by $X$.
Generally, one can easily construct multimodal distributions for which the expectation might lie somewhere between the modes in an arbitrary large area of $0$ probability.
\end{example}
\end{tcolorbox}
\end{figure}

As the expectation is describing some aspect of the distribution over $\X$, an immediate idea is to also consider expectations of transformations of $\X$.
In particular, for some (measurable) function $g$ and we might define $\Y = g(\X)$.
Then, the expectation of $\E[\Y]$ will also be describing aspects of the original distribution $p_{\X}$.
Computing $\E[\Y]$ in principle requires that we first figure out $p_{\Y}$, which is often a delicate task.
However, there is another and often simpler way, as it holds that 
\begin{equation}
\E[g(\X)] = \E[\Y] = \int p(\y) \y \, \mathrm{d}\y = \int p(\x) g(\x) \, \mathrm{d}\x.
\end{equation}
This identity is known as the \emph{law of the unconscious statistician (LOTUS)}, as the right hand side is often considered as a definition of $E[g(\X)]$, even though it is actually a theorem.

For univariate random variables, a common choice for $g$ are powers, i.e.~$g(x) = x^K$, where $k \in \naturals$, and the expectation $\E[X^k]$ is called the \emph{$k$\textsuperscript{th} moment} of $p_{X}$.
The first moment is of course just the expectation $\E[X]$ also called the \emph{mean} of $p_X$.
The second moment of a transformed version of $X$, centered at 0 by subtracting its mean, is the \emph{variance}
\begin{equation}
\mathds{V}[X] \coloneqq \E[(X - \E[X])^2] = \E[X^2] - \E[X]^2.
\end{equation}
For random vectors $\X$, the corresponding concept is the \emph{covariance matrix}:
\begin{equation}
Cov[\X] \coloneqq \E[(\X - \E[X])^T (\X - \E[X])] = \E[\X^T \X] - \E[\X]^T \E[\X].
\end{equation}
The diagonal of $Cov$ are the variances of the individual random variables in $\X$, while the off-diagonal contains the covariances between any pair $X,Y \in \X, X \not= Y$:
\begin{equation}
Cov[X,Y] = \E \left[(X - \E[X]) \, (Y - \E[Y]) \right].
\end{equation}

In summary, expectations are one of the main tool to convert probability distributions into ``concrete values''.
This concept becomes in particular important when paired with optimization (taking maxima or minima), leading to \emph{decision making}, see Section~\ref{sec:decision_theory}.

\section{Most Probable Explanation}

Rather than averaging we might also take the extreme values, i.e.~maxima or minima.
In particular, when $p$ is a joint distribution over random variables $\X = (X_1, \dots, X_D)$ with state space $\mathbfcal{X}$, then we might ask which state is the most likely one, i.e.~the state with maximal probability:
\begin{equation}
\x^* = \arg\max_{\x \in \mathbfcal{X}} p(\x)
\end{equation}
Here we assume the maximum is unique.
If it is not, we might either break ties in some way or return the set of all maxima.
The value $\x^*$ is known as \emph{most probable explanation (MPE)} \cite{pearl1988probabilistic}.

The MPE is particularly interesting for conditional distributions.
Referring to Example~\ref{ex:blood_flow}, the MPE for $Y$ given the observed information would be 
\begin{equation}
y^* = \arg\max_{y} p(y \cbar x_1,\dots,x_{16},x_{18},x_{20})
\end{equation}
If $Y$ is a discrete random variable, and hence the conditional distribution function is a PMF, $y^*$ is indeed the most probable value for $Y$ for the given evidence.

\begin{figure}
\begin{tcolorbox}[width=\textwidth,]
\begin{example}[MPE with Densities]
\label{ex:mpe_density}
Assume that $Y$ has the following density
$$
p_Y(y) = 0.999 p_{{N}}(y \cbar -10, 2) + 0.001 p_{{N}}(y \cbar 9, 0.0005), 
$$
where $p_{{N}}(y \cbar \mu, \sigma) = {1 \over \sqrt{2 \pi \sigma^2}} e^{- {(x - \mu)^2 \over 2\sigma^2}}$ is the Gaussian density.
Hence, $p(y)$ is a \emph{Gaussian mixture} with two components, one of which represents $0.999$ of the total probability mass and the other $0.001$.  
This density is shown in the following figure:
\halfbline
			
\begin{center}
\includegraphics[width=0.85\textwidth]{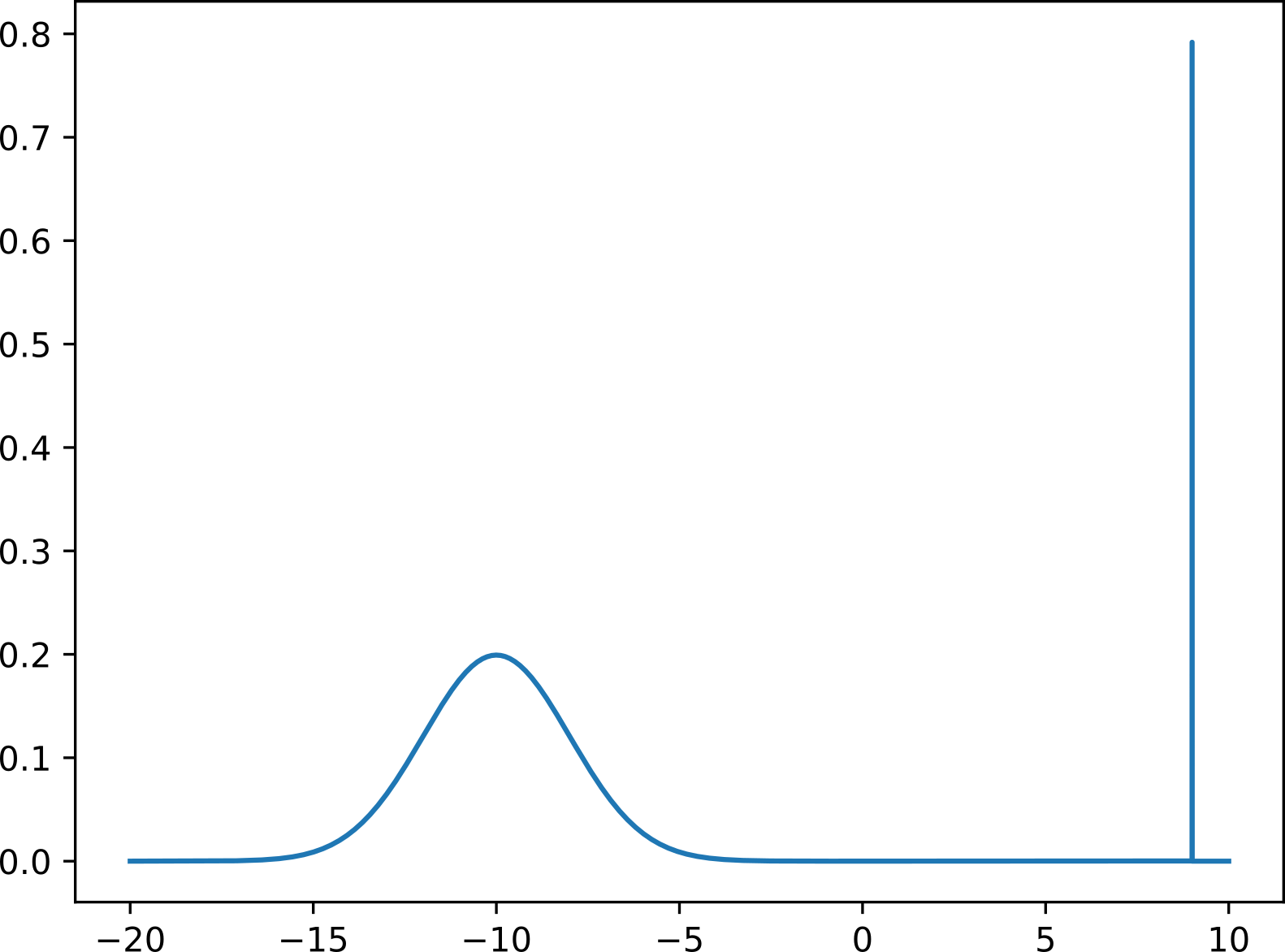}
\end{center}

Here we omit any conditioning information, but we might figure $p_Y$ as a (toy) distribution resulting from \eqref{eq:infer_bloodflow} in Example~\ref{ex:blood_flow}.
The MPE in this example, i.e.~the state where $p_Y$ is maximal, is $y^*=9$,  or actually slightly less, due to the Gaussian component on the left.
\end{example}
\end{tcolorbox}
\end{figure}

When $Y$ is continuous, and hence the conditional is a density, the interpretation of $y^*$ is a bit more subtle, as illustrated in Example~\ref{ex:mpe_density}.
The MPE in this example is perhaps surprising, as the value $Y=9$ maximizes the density $p_Y(9) \approx 0.8$, but is located in a region attributing only about $0.1\%$ of the total probability mass.
If we were to pick the other mode around $Y=-10$, we would get a smaller value for $p_Y(-10) \approx 0.2$ but be in a region accounting for $99.9\%$ of the probability.

Is MPE flawed for densities?
No, the seeming paradox comes from making tacit assumptions.
MPE makes the following promise: assuming that $p_Y$ is {continuous}, then there exists some $\epsilon > 0$ such that the probability that $Y$ falls into the interval $[y^* - \epsilon, y^* + \epsilon]$ is maximal when $y^*$ is the MPE.
In Example~\ref{ex:mpe_density}, if we discretize the state space of $Y$ into very small cells and bet on the outcome in which cell $Y$ will land, we are best off by committing to the MPE at $y^*=9$ (for small enough cell size).
This might often not be what we want for real-valued state spaces.
The solution to such apparent paradoxes is offered by decision theory.

\section{Decision Theory}
\label{sec:decision_theory}

The issue with Example~\ref{ex:mpe_density} is that many would prefer the mode of the left Gaussian, since with $99.9\%$ probability $Y$ takes a value in the vicinity of $-10$.
Intuitively, if we guess $Y=-10$ but the true value turns out to be, say, $Y=-10.13$, we still would be happy, since we have guessed a value \emph{close} to the true value.
As illustrated above, MPE essentially discretizes the space into tiny cells and deliberately ignores any notion of \textit{closeness} among the cells---if we bet on the neighboring cell of where $Y$ really lands, ``we still lose'' according to MPE.
Decision theory \cite{berger1990statistical} allows us to make such assumptions explicit, by introducing a \emph{loss function} and a simple recipe for making optimal decisions under uncertainty.

Assume we aim to decide on the value of a scalar quantity of interest $Y$ (the case for random vectors is straightforward), and let  $\mathcal{Y}$ be the state space of $Y$.
In order to make this decision we assume that we have evidence $\X=\x$, and that we have already computed the conditional $p_{Y|\X}$, by applying the sum and product rule.
If no evidence is available, we would use the marginal $p_{Y}$ instead.
In either case, we arrive at the required distribution by mere application of the sum and product rules.

A \emph{loss function} $\ell$ is any function
\begin{equation}
\ell\colon \mathcal{Y} \times \mathcal{Y} \mapsto \reals.
\end{equation}
Given $y^*,y_{t} \in \mathcal{Y}$, the value $\ell(y^*, y_{t})$ represents our regret when we decide for value $y^*$, but actually $Y=y_{t}$.
Usually, loss functions will satisfy $\ell(y_{t},y_{t}) \leq \ell(y^*,y_{t})$ for all $y^* \in \mathcal{Y}$, that is, our loss will be minimal if we correctly guess the true value of $Y$.

Now, the basic rule of decision theory is to decide for the value which minimizes the \emph{expected loss}, conditional on the available evidence $\X=\x$:
\begin{equation}
\label{eq:minimal_expected_loss}
y^* = \arg\min_{y \in \mathcal{Y}} ~ \E[\ell(y, Y) \cbar \x].
\end{equation}
The frequentist interpretation of probability gives a clear argument why this is an optimal decision rule: in a scenario of making repeated decisions, rule \ref{eq:minimal_expected_loss} will lead to the minimal sum of losses if we correctly inferred the true $p_{Y|\X}$ and the number of decision trials goes to infinity.
Let us study a few basic loss functions in the context of Example~\ref{ex:mpe_density}.

\paragraph{Zero-one loss.}
Consider for some $\epsilon > 0$ the loss function
$$
\ell_{0/1}(y^*, y) = 
\begin{cases}
0 & \text{if } |y^*-y| < \epsilon \\
1 & \text{otherwise},
\end{cases}
$$
denoted as \emph{zero-one} loss. 
If the (conditional) density of $Y$ is continuous, then \eqref{eq:minimal_expected_loss} under zero-one loss approaches the MPE when $\epsilon \rightarrow 0$.
Hence MPE essentially minimizes zero-one loss.

\paragraph{Squared distance loss ($\ell_2$-loss).}
The ``all-or-nothing'' behavior of MPE is probably not what one wants in Example ~\ref{ex:mpe_density}, as it delivers an ``atypical'' value in an area of low probability.
Rather, one would probably prefer the \emph{squared distance loss}
$$
\ell_2(y^*, y) = (y^* - y)^2,
$$
as it assumes that we are still happy if we guess a values close to the true value of $Y$.
It can be shown that for any distribution $p_{Y}$, the \emph{least squares loss}, i.e.~\eqref{eq:minimal_expected_loss} when using $\ell_2$, is achieved by the mean of $p_Y$.
For a Gaussian mixture model the mean is given as the corresponding mixture of the components' means. 
Specifically, in Example~\ref{ex:mpe_density}, we get the solution
\begin{equation}
\label{eq:squared_loss}
y^* = 0.999 \times (-10) + 0.001 \times 9 = -9.981,
\end{equation}
delivering a value almost at the center of the high probability region on the left.

\paragraph{Absolute distance loss ($\ell_1$-loss).}
However, one might be concerned by the fact that the least squares solution \eqref{eq:squared_loss} is quite influenced by the locations of the Gaussian components.
In particular, if the mean of the right Gaussian component (the one with weight $0.001$) was moved to a very large value, e.g.~$9000$, we would get $y^*=-0.99$ which is far away from the high probability area.
This proneness to outliers is typical for least squares.
A remedy to this problem is to use \emph{absolute distance loss} defined as
$$
\ell_1(y^*, y) = |y^* - y|.
$$
It can be shown that the minimum of \eqref{eq:minimal_expected_loss} when using $\ell_1$ is achieved by the \emph{median} of $p_Y$, which is the value $y^*$ for which $\prob(Y \leq y^*) = \prob(Y \geq y^*) = 0.5$.  
It can be figured out that the median in Example \ref{ex:mpe_density} is about $-9.9975$ and it stays virtually the same if we move the mean of the right Gaussian to $9000$ or even far more extreme values.
This robustness to outliers is a well known property of $\ell_1$-loss and the median.

\vspace{\baselineskip}

In summary, minimizing expected loss \eqref{eq:minimal_expected_loss} is a principled approach to ``call the shots'' and transit from our probabilistic prediction---a conditional distribution over the quantity of interest $Y$---to a concrete hard value $y^*$. 
It should be advised that---in the ideal setting---this step should come very last, since it discards all the beautiful ``fluffy uncertainty'' about $Y$ and approximates it with a brutally hard single value.

In particular in Example~\ref{ex:blood_flow}, figure the scenario that inferring the cerebral blood flow $Y$ was only an intermediate step, and that the ultimate goal is to infer, together with other medical parameters, the health condition of a patient.
In this scenario it is crucial to keep the entire $p(y \cbar \dots)$ and process it further downstream.
Approximating it in any way before we have arrived at the final inference goal compromises the exact reasoning calculus of probability, destroys valuable information, and can---in principle---deliver arbitrary bad results.
In other words, we shall ``stay calm'' and continue applying the sum and product rule.\footnote{A typical criticism raised at this point is that probabilistic modeling and inference are hard, and that all the required approximations along the way might be much worse than replacing intermediate distributions with concrete values. This is certainly a valid point, but conflates two aspects: one aspect, which is the central one in this thesis, is asking what \textbf{in principle} should be done, i.e.~what is in principle an optimal calculus of reasoning under uncertainty? The other aspect is what can be done in reality and how well our optimal calculus can actually be implemented.}

Some points are worth mentioning.
First, defining a loss function $\ell$ and applying decision rule \eqref{eq:minimal_expected_loss} makes our assumptions \emph{explicit} and allows us to specify about what we care when deciding a value $y^*$.
However, designing a loss function for a concrete application is not trivial.
For example, when we are dealing with a \emph{binary classification} problem, i.e.~$Y \in \{0,1\}$, where $Y=0$ represents the absence of a deadly disease and $Y=1$ represents its presence, it is quite delicate to come up with good loss function.   
It is clear, that predicting $y^*=0$ (patient healthy) when in fact $Y=1$ (patient ill) should incur a higher loss than for the converse case.  
But, how much higher? 
We might fix an arbitrary price, e.g.~$1$ Euro, to diagnosing the patient as ill when in fact they are healthy, but how much does it cost when we release an actually ill patient and endanger their live?
An ethically delicate question.

Yet, I want to stress again that \emph{if} (i) the loss function $\ell$ indeed adequately represents our preferences, (ii) we have access to the true data generating distribution $p_{Y,\X}$ and (iii) we can actually compute  \eqref{eq:minimal_expected_loss} \emph{exactly}, there is nothing left to do---we have successfully made an optimal decision in terms of expected loss.
There are a lot of ``if's'' here and I am not saying that these computations are easy, yet it is surely beneficial to know and specify what is the best one can do.  
For example, in the case of classification, the MPE computed for the true data distribution leads to the \emph{Bayes optimal classifier}, i.e.~the classifier with minimal zero-one loss.
A famous result for the \emph{K-nearest neighbor classifier} is that it asymptotically converges to the Bayes optimal classifier (given that the number of neighbors grows also to infinity, but less strongly than the number of data points), making it a theoretically well-justified classification rule \cite{bishop2006pattern}.
In the context of regression, using the data generating distribution and $\ell_2$ loss leads to the \emph{optimal regression function}, and no ever-so-fancy machine learning method will be able to outperform it in terms of true squared loss.

Finally, it is worth mentioning that \eqref{eq:minimal_expected_loss} is not the only good decision rule.
For example, we might not care about the \emph{expected loss} too much but rather predict a value such that the suffered loss is smaller than a certain threshold with a prescribed probability, and otherwise reject the decision, leading to a more risk aware prediction scheme.
Moreover, \emph{conformal prediction} \cite{shafer2008tutorial} gives up on the property to return only a single value but rather returns a minimal set of predictions $\mathcal{Y}^* \subset \mathcal{Y}$ such that the true value of $Y$ ends up in in $\mathcal{Y}^*$ with high probability.

%% file: ChapterSOTA/chapterSOTA.tex
\chapter{State of The Art}   
\label{chap:SOTA}

In this chapter, I provide an overview of state-of-the-art probabilistic approaches in AI.  
While AI was largely dominated by logic-based methods until the late 1980s, the necessity of accounting for uncertainty in real-world scenarios became increasingly recognized.  
Although probability was not immediately embraced---it actually faced quite some criticism for its computational challenges and perceived impracticality---it gradually gained ground. A key proponent was Judea Pearl, whose seminal book \textit{Probabilistic Reasoning in Intelligent Systems: Networks of Plausible Inference} \cite{pearl1988probabilistic} firmly established probabilistic reasoning as a central approach in AI.

\paragraph{Graphical Models,}
in particular Bayesian networks, were central to Pearl’s work and have since remained a cornerstone of probabilistic modeling \cite{koller2009probabilistic}. 
Although their prominence has somewhat diminished in the deep learning era, they still serve as a \textit{lingua franca} of probabilistic modeling, due to their intuitive visual representation, which correspond directly with conditional independence assumptions and tractability in inference and learning \cite{pearl1988probabilistic,koller2009probabilistic}.  

Inference in graphical models---especially marginalization via the sum rule---is generally NP-hard \cite{cooper1990computational}. However, exact inference is tractable for models with bounded \textit{treewidth}, a measure of how tree-like a graph is. Most probabilistic machine learning systems can be mapped onto a corresponding graphical model, and authors often use such them to concisely express model structure and assumptions.

\subparagraph{Deep Autoregressive Models}
\cite{larochelle2011neural,van2016conditional} apply the chain rule of probability and model each conditional distribution using a shared neural network. These models can be interpreted as fully connected Bayesian networks, which are capable, in principle, of representing any joint distribution.  
While they support sampling and density evaluation, computing marginals, conditionals, or performing complex inference tasks remains challenging.

\subparagraph{Variational Autoencoders (VAEs)}
\cite{kingma2019introduction} are a neural latent variable model, previously known as \emph{density models} \cite{mackay1995bayesian}.
They introduce a latent code with a simple distribution, often isotropic Gaussian, which is transformed by a neural network to produce the (parameters of) a conditional distribution over observable variables; hence they represent an infinite mixture model, with the latent variable marginalized out.

Exact inference in VAEs (density networks) is a hard task, but the \emph{evidence lower bound} (ELBO), that is, a lower bound on the log-likelihood can be estimated.
A major innovation of VAEs was the introduction of \textit{amortized inference}: instead of computing the posterior for each data point individually, a separate network---commonly called the \textit{encoder} or \textit{inference network}---learns to approximate it.  
The encoder is used to compute an ELBO estimate which used as the VAEs learning objective as a replacement for the exact log-likelihood.
By optimizing both the parameters of the model and the encoder, one achieves the dual goal to improve the ELBO estimate and optimizing the model parameters at the same time.

\subparagraph{Generative Adversarial Networks (GANs)} \cite{goodfellow2014generative} also introduce a simple latent distribution but adopt an adversarial training scheme rather than maximizing a log-likelihood bound. 
They involve a second network---the \textit{discriminator} or \textit{critic}---which learns to distinguish real data from generated samples.  
This training objective corresponds to minimizing a probabilistic divergence distinct from maximum likelihood \cite{goodfellow2014generative,arjovsky2017wasserstein}.  

While GANs enable realistic sampling, they generally do not allow density evaluation (often such densities do not exist in the Lebesgue sense), nor do they support tractable inference for marginals or conditionals.

\subparagraph{Normalizing Flows}
\cite{rezende2015variational,papamakarios2021normalizing}
are another principle to probabilistic modeling.
Similar as VAEs and GANs, they start with a simple distribution (e.g., an isotropic Gaussian) and apply a series of bijective, differentiable transformations to generate samples from a more complex target distribution.  
The exact density of the transformed distribution is computed via the change of variables formula, and exact sampling is also feasible. 
However, computing marginals or conditionals remains challenging, similar to VAEs and GANs.

\subparagraph{Energy-Based Models} define distributions using an unnormalized energy function \( E(\mathbf{x}) \), often parameterized by a neural network:
\[
p(\mathbf{x}) = \frac{\exp(-E(\mathbf{x}))}{\int \exp(-E(\mathbf{x})) \mathrm{d}\mathbf{x}}.
\]
While extremely flexible, these models are typically intractable. Evaluating densities, drawing samples, and performing inference are all computationally expensive. Marginals and conditionals are especially difficult to estimate reliably.

\subparagraph{Score-Based Models} \cite{vincent2011connection} are closely related to energy-based models and represent a distribution via its \textit{score function}, i.e., the gradient of the log-density, \( \nabla \log p(\mathbf{x}) \). 
A neural network is trained to approximate this score.  
When combined with diffusion processes and Langevin dynamics, these models currently represent the state of the art in generative modeling for images \cite{song2020score}.  

Despite their impressive performance, score-based models usually lack tractable density evaluation and require approximations for sampling, marginals, and conditionals. 
These approximations can be difficult to validate in practice.

\subparagraph{Gaussian Processes (GPs)} \cite{williams2006gaussian} are another prominent type of probabilistic model.  
Formally, GPs are a (possibly uncountable) collection of random variables, of which any finite subset has a multivariate Gaussian distribution.  
Usually, these random variables are indexed by some continuous parameter, e.g.~$x \in \mathbb{R}$, which allows us to interpret GPs as random functions.  
GPs have ample applications in Bayesian modeling of latent functions and Bayesian optimization \cite{snoek2012practical}.  
Due to Gaussianity, computing marginals and conditionals in GPs is tractable, but associated with a cubic computational cost in the number of data points, which becomes delicate for large data sets.  
These challenges are most commonly addressed with \emph{sparse inducing point techniques} \cite{quinonero2005unifying}.

\subparagraph{Probabilistic Programming (PP)} \cite{van2018introduction}  
is a general framework to express probabilistic models in a flexible way, namely by writing a program equipped with random variables.  
When evaluating such a program in the standard way, it simply amounts to a stochastic simulator.  
However, the main powerful idea of PP is that such a program constitutes a probabilistic model which can be subject to inference.  
In this way, one might provide data for the observed output of a probabilistic program and query the posterior distribution over certain latent parameters, in principle inferred via Bayes' rule.  
In practice, this inference is tackled via approximation, usually employing variational inference or Monte Carlo techniques.  
In a nutshell, PP aims to achieve for probabilistic modeling and inference what machine learning frameworks like TensorFlow and PyTorch have achieved for deep learning.

\bline

The list of probabilistic modeling approaches above is not complete, but captures a coarse picture of ``what is going on in the field''.
A remarkable point is that many models mainly strive for expressivity and neglect the question about inference, or actually, heavily rely on approximation such as sampling and variational inference.\footnote{The most noteworthy exceptions are low-tree width graphical models and GPs.}
This is in quite stark contrast to the picture of probability as a sound and rigorous reasoning framework under uncertainty.
Reasons for this neglect are, in my personal opinion, stemming from a particular mindset (i) assuming that tractable models necessarily underperform and (ii) tailored towards the contemporary success of deep learning (i.e.~function approximation) style of AI.
Probabilistic circuits, as discussed in the following, put an explicit focus on tractability.

%% file: Chapter4/chapter4.tex
\chapter{Probabilistic Circuits}   
\label{chap:probabilistic_circuits}

So far, I have presented probability as an elegant and rigorous inference language for AI, boiling down to two simple rules, the sum and product rule.
Then, in order to make (optimal) decisions, all we need to do on top of that is taking expectations, maxima and minima.
Of course, this actually requires us that we have access to the \emph{true} underlying data distribution or a reasonably good approximation.
If our model defers too much from reality, all the inferences we are doing will be worthless---garbage in, garbage out. 
Hence, faithful modeling and learning is a crucial part of the game.

Probabilistic models are generally some representation of a distribution function $p_{\X}$, such as \emph{probabilistic graphical models} (PGMs) \cite{koller2009probabilistic}, \emph{density networks} and \emph{variational autoencoders} (DNs, VAEs) \cite{mackay1995bayesian,kingma2019introduction,rezende2014stochastic}, \emph{neural auto-regressive distribution estimators} (NADEs) \cite{larochelle2011neural}, \emph{energy-based models} \cite{song2021train}, \emph{score-based models} \cite{vincent2011connection}, 
\emph{normalizing flows} \cite{rezende2015variational}, \emph{probabilistic programs} \cite{bingham2019pyro}, and \emph{Gaussian processes} (GPs) \cite{williams2006gaussian}.\footnote{GPs are technically a collection of (infinitely many) random variables with the property that any finite dimensional marginal is Gaussian. Hence, the distribution function of any finite subset of random variables is defined.}
A common model which does not represent a distribution function $p_{\X}$ (yet a distribution $\prob_{\X}$) is \emph{generative adversarial networks} (GANs) \cite{goodfellow2014generative}, as they usually do not admit a density with respect to Lebesgue measure.
All these models represent distributions in various creative and flexible ways, so that we are hopefully able to represent and learn the distribution we are interested in in.

Probabilistic inference, however, is generally intractable in most of these models.
Exceptions are PGMs with \emph{bounded tree-width} \cite{koller2009probabilistic} (computing marginals is exponential in the tree-width) and GPs (computing conditionals is cubic in the number of samples).
For all other models, computing marginals and/or conditionals is a hard computational problem, as they are all based on high-dimensional non-linear functions (neural networks).
Computing integrals---the core business of inference---over such functions is a notoriously hard task.

\emph{Probabilistic circuits (PCs} \cite{vergari2020probabilistic,choi2020probabilistic}
are a type of \textbf{tractable} probabilistic model that allows to perform inference \textbf{exactly} and \textbf{efficiently}, meaning that the probabilistic inference task of choice (marginals, conditionals, etc.) are computable without approximation\footnote{Leaving aside numerical errors stemming from finite precision arithmetic.} and in polynomial time of the PC's size.
Hence, PCs stay at all times faithful to the probabilistic reasoning principle: when we manage to represent our distribution of interest as a PC, we do have the guarantee that we arrive at the correct conclusion.
This property makes PCs quite unique among other probabilistic models, which typically strive for maximal expressivity rather than tractable inference.

Evidently, there must be a catch to PCs. 
Basically, PCs ``buy'' tractability by accepting rather stringent constraints on (or properties of) their structure.
These constraints make them less expressive than other types of model, meaning that there exist probability distributions which have a compact representation (polynomial in the number of random variables) as, say, a PGM, but an exponential representation size as a PC. 
These theoretical arguments, which go under the name \emph{expressive efficiency} \cite{martens2014expressive}, are however not fully conclusive.
In particular, the PC community has steadily been making progress over the last two decades and demonstrated that tractable models often compete well with intractable ones or even outperform them. 
Some examples are provided in Part II of this thesis.\footnote{For more examples, see e.g.~our latest tutorial at NeurIPS'22 \cite{vergari2022probabilistic}.}
Furthermore, it should be noted that PCs are still universal approximators of densities, that is, they absolutely \emph{can} represent any distribution we like.
The representation size, however, might just be infeasible.

The term ``probabilistic circuits'' is a relatively new one, but the principle has been around since decades and is known under many different names.
\emph{Arithmetic circuits} \cite{darwiche2003differential} are probably the earliest representatives and clearly display the key ideas of PCs and other tractable circuits (operating e.g.~in the logic domain).
The concept of PCs is also reflected in \emph{AND/OR-graphs} \cite{dechter2007and}, \emph{sum-product networks} \cite{poon2011sum} and many other models.
The main idea to use yet another name, put forward by several researchers at UCLA \cite{choi2020probabilistic}, is to unify these superficially different names and concepts under one umbrella.

In this chapter, I introduce PC and explain their working principles.
Furthermore, I give an overview of the work I have done in this area over the last 10 years, which also concludes Part I of this cumulative Habilitation thesis.
In Part II, the relevant papers I have published are listed as chapters.

\section{Basic Definitions of Probabilistic Circuits}   
\label{sec:basic_definitions_pcs}

We start with the formal definition of PCs.

\begin{definition}[Probabilistic Circuit (PC)]   
\label{def:probabilistic_circuit}

Let a collection of random variables $\X = (X_1, \dots, X_D)$ be given, let $\XS_1, \dots, \XS_D$ be their state spaces and $\mathbfcal{X} = \bigtimes_{i} \XS_i$ the joint state space of $\X$.
A \emph{probabilistic circuit (PC)} is a representation (model) of a potentially unnormalized joint probability distribution defined on $\mathbfcal{X}$, i.e., a function 
\begin{equation}
\label{eq:PC_function_signature}
p\colon \mathbfcal{X} \mapsto [0, \infty]. 
\end{equation}

Specifically, a PC is a type of neural network, i.e.~a \emph{(directed acyclic) computational graph} $\graph = (V, E)$ containing three types of \emph{computational nodes}, namely \emph{probability distribution functions} (short \emph{distributions}), \emph{weighted sums} and \emph{products}.
In the following we introduce the symbols $\distnode$, $\sumnode$ and $\prodnode$ to refer to some generic distribution, sum and product node, respectively, while $\node$ refers to a generic node of any type.
Further, let $\pmb{\distnode}$, $\pmb{\sumnode}$ and $\pmb{\prodnode}$ denote the set of all distributions, sums and products, respectively.

For any $\node \in V$, let $\inn(\node)$ denote the neighbors of $\node$ with an edge towards $\node$, and $\outn(\node)$ be the neighbors with an edge outgoing from $\node$.   
The input to the PC is some state $\x = (x_1, \dots, x_D) \in \mathbfcal{X}$ for the random variables ($\x$ is not contained in $V$).

The leaves of $\graph$ are the distribution nodes, i.e., {$\pmb{\distnode} = \{\node \in V \cbar \inn(\node) = \emptyset\}$}, while all other nodes are either sum or product nodes.
Moreover, only the distribution nodes get $\x$ as direct input (and by construction no input from any $\node \in V$), while sum and product nodes can only get direct input from nodes in $V$.
Thus, the distribution nodes $\pmb{\distnode}$ form the first layer in the PC, while sums and products are located on ``higher'' layers.
For that reason, distribution nodes are also often called \emph{input distributions} of the PC.

A \textbf{distribution node} $\distnode$ is a distribution function over some subset of $\X$.
This subset, written as $\X[\distnode] \subseteq \X$, is denoted as the \emph{scope} or \emph{receptive field} of $\distnode$.
Hence, when evaluating the PC for some joint state $\x$, node $\distnode$ only receives $\x[\distnode]$, defined as the vector containing the values in $\x$ corresponding to $\X[\distnode]$.
The value computed by $\distnode$ is the distribution function evaluated at $\x[\distnode]$, which is technically $\distnode(\x[\distnode])$, but usually written simply as $\distnode(\x)$.
A distribution node might have trainable parameters, which are denoted as $\bm{\theta}_{\distnode}$.

A \textbf{sum node} $\sumnode$ with real weights ${\bm{\theta}}_{\sumnode}=(\theta_{\sumnode\node})_{\node \in \inn(\sumnode)}$ computes a nonnegative mixture of its inputs, i.e., 
\begin{equation}
\label{eq:sumnode}
\sumnode(\x) = \sum_{\node \in \inn(\sumnode)} \theta_{\sumnode\node} \, \node(\x)
\end{equation}
where $\theta_{\sumnode\node}$ are (trainable) weights satisfying $\theta_{\sumnode\node} \geq 0$.

A \textbf{product node} $\prodnode$ computes the product of its inputs, i.e., 
\begin{equation}
\label{eq:prodnode}
\prodnode(\x) = \prod_{\node \in \inn(\prodnode)} \node(\x).
\end{equation}

Note since distribution nodes have a restricted scope, that is, a subset of $\X$, sum and product nodes will also have a restricted scope.
For an arbitrary sum or product node $\node$, its scope is recursively given as
\begin{equation}
\label{eq:recursive_scope}
\X[\node] = \bigcup_{\node' \in \inn(\node)} \X[\node'].
\end{equation}

The nodes with $\outn(\node) = \emptyset$ are called the roots or output nodes of the PC.
We assume that the roots have full scope, i.e., $\X[\node] = \X$.
Usually, we assume that the PC has a single root, representing the model \eqref{eq:PC_function_signature}.
\end{definition}

\begin{figure}
\begin{tcolorbox}[width=\textwidth,]
\begin{example}[Probabilistic Circuit]
\label{ex:example_pc}
~\\~\\
\begin{minipage}{0.55\textwidth}
An example PC over seven random variables $X_1, \dots, X_7$ is shown on the right.
A complete state (observation) $\x = (x_1,\dots,x_7)$ serves as input to the PC.
The first layer consists of distributions nodes ($\distnode$'s), here illustrated with nodes showing a Gaussian-like distribution functions (they are, however, not necessarily Gaussian, but any distribution function can be used).
The scope of each distribution node can be recognized by which input values are connected to it.  

The nodes with $+$ and $\times$ symbols are the sum and product nodes, respectively.
The PC computes a density by evaluating the circuit bottom up, as usual in neural networks.
The return value is $p(\x)$, the PC distribution evaluated at $\x$. 
Paramters for sum weights and distribution nodes are not shown in the figure.
\end{minipage}~~~~~~~
\begin{minipage}{0.42\textwidth}
\includegraphics[width=0.99\textwidth]{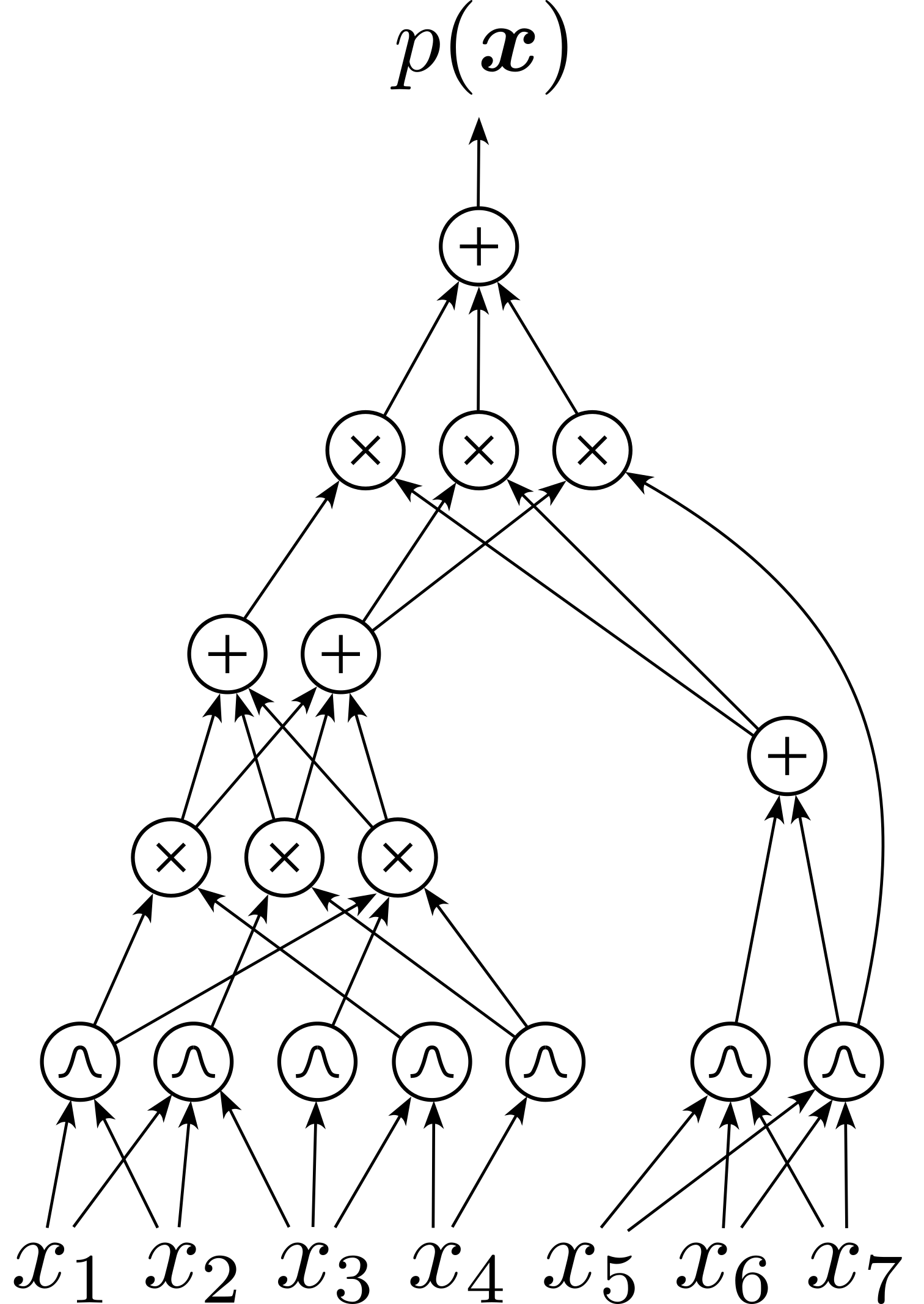}
\end{minipage}
\end{example}
\end{tcolorbox}
\end{figure}

We see that the definition of PCs is quite straightforward:
We aim to represent a multivariate distribution function $p\colon \mathbfcal{X} \mapsto [0, \infty]$ and do this with a neural network whose first layer consists of distribution functions over smaller scopes, and whose higher layers contains nonnegatively weighted sums and products.
Since the distribution nodes are nonnegative and the sum weights are nonnegative, the whole function will be nonnegative.
Of course, PCs according to Definition~\ref{def:probabilistic_circuit} might not be normalized, hence they define a distribution function up to the normalization constant
\begin{equation}
\label{eq:pc_z}
\mathcal{Z} = \int p(\x) \, \mathrm{d}\x,
\end{equation}
which we assume to be finite and non-zero.
An example of a PC is shown on Example~\ref{ex:example_pc}.

\subsection{Decomposability and Smoothness}

The central motivation of PCs is that probabilistic inference remains tractable, i.e., so at least we should be able to compute marginals and conditionals.
However, PCs according to Definition~\ref{def:probabilistic_circuit} are actually intractable.
In particular, the normalization constant \eqref{eq:pc_z} and any other marginals will be pretty hard to compute.
We are still missing some ``secret sauce'' for tractability, namely, \emph{structural constraints}. 
The two most important structural constrains used in PCs are \emph{decomposability} and \emph{smoothness}.

\begin{definition}[Decomposability]
A product node $\prodnode$ in a PC is called \emph{decomposable} if its input nodes have \emph{disjoint} scopes, i.e.,
$$
\forall \node, \node' \in \inn(\prodnode), \node \not= \node' \colon \X[\node] \cap \X[\node'] = \emptyset. 
$$
A PC is decomposable if all its product nodes are decomposable.
\end{definition}

\begin{definition}[Smoothness]
A sum node $\sumnode$ in a PC is called \emph{smooth} if all its input nodes have the \emph{same} scope, i.e.,
$$
\forall \node,\node' \in \inn(\sumnode) \colon \X[\node] = \X[\node']. 
$$
Note that by \eqref{eq:recursive_scope} a smooth sum node has the same scope as any of its input nodes.
A PC is smooth if all its sum nodes are smooth.
\end{definition}

Note that the PC in Example~\ref{ex:example_pc} is decomposable and smooth.
Smoothness is a bit of a cosmetic property, as a non-smooth PC can either be rendered smooth or it turns out that the normalization constant is $\infty$, hence the PC is not a valid representation of a distribution.
Hence, for now we can just safely assume that any PC is smooth.
Decomposability will turn out to be a key property for tractable inference, in particular marginals and conditionals.
But first let us better understand what these two structural properties imply.

\subsection{PCs as Hierarchical Mixtures}
\label{sec:hierarchical_mixtures}

An immediate and convenient consequence of decomposability and smoothness is that the normalization constant of a PC is $1$ if we (i) use normalized input distributions (i.e.~their normalization constant is $1$) and (ii) use normalized sum weights, i.e., for each $\sumnode$, additionally to requiring $\theta_{\sumnode\node} \geq 0$ we also require $\sum_{\node \in \inn(\sumnode)} \theta_{\sumnode\node} = 1$, so that $\sumnode$ computes a \emph{convex combination} of its inputs.
More precisely, it will turn out that \emph{each} node in a PC is a normalized distribution over its scope and that the whole PC just forms a hierarchical mixture model.

We show this property by induction over a \emph{topological order} of the PC nodes $V$.
Since the computational graph is acyclic and directed, there exists at least one topological order of the nodes
\begin{equation}
\label{eq:topological_order}
\node_1 \prec \node_2 \prec \dots \prec \node_{|V|}, 
\end{equation}
such that it holds for each $\node_i$ that all nodes in $\inn(\node_i)$ have an index smaller than $i$.
In other words, order \eqref{eq:topological_order} is a topological order if arrows can only point from ``left to right''.
The existence of at least one topological order is equivalent to the fact that the PC graph $\graph$ is directed and acyclic.

We can assume that the input distributions are the first $|\pmb{\distnode}|$ nodes in this order.
Since we assume that they are already properly normalized distributions, their normalization constant is $1$.
This is the induction basis.

Next assume any node $\node_i$ with $i > |\pmb{\distnode}|$, which is either a sum or product node.
We assume, by induction, that all nodes $\node_1,\dots,\node_{i-1}$ have normalization constant $1$ and show that in this case also $\node_i$ has normalization constant $1$.
First note that all $\inn(\node_i)$ come before $\node_i$ in the order and have therefore a normalization constant of $1$.

Assume first that $\node_i = \prodnode$ is a product.
For simplicity, we assume that $\prodnode$ has only two inputs $\node$ and $\node'$.
The case for more inputs is immediate.
Further, let us rename the scopes of the two input nodes as $\X[\node] = \Y$ and $\X[\node'] = \Z$ and recall that decomposability means that $\Y \cap \Z = \emptyset$.
Then the product computes
\begin{equation}
\label{eq:product_node_factorization}
\prodnode(\x) = \node(\y) \, \node'(\z).
\end{equation}
Hence, since $\node$ and $\node'$ are correctly normalized distributions, $\prodnode$ is just a \emph{factorized} distribution, assuming independence among $\Y$ and $\Z$, see Section~\ref{sec:conditional_independence_distributions}.
In particular, its normalization constant is $1$ since
\begin{equation}
\int \int \prodnode(\x) \, \mathrm{d}\y \,\mathrm{d}\z  
=
\int \int \node(\y) \, \node'(\z) \, \mathrm{d}\y \,\mathrm{d}\z  
=
\overbrace{\left(\int \node(\y) \, \mathrm{d}\y \right)}^{=1} 
\, \overbrace{\left( \int \node'(\z) \,\mathrm{d}\z \right)}^{=1} = 1.
\end{equation}

If, on the other hand, $\node_i$ is a sum node, it computes 
\begin{equation}
\label{eq:sum_mixture}
\sumnode(\x) = \sum_{\node \in \inn(\sumnode)} \theta_{\sumnode\node} \node(\x).
\end{equation}
Since by induction hypothesis all $\inn(\sumnode)$ are correctly normalized distributions, and since the weights satisfy $\theta_{\sumnode\node} \geq 0$ and $\sum_{\node \in \inn(\sumnode)} \theta_{\sumnode\node} = 1$ and the sum node is smooth, 
\eqref{eq:sum_mixture} is just a \emph{mixture distribution}, i.e.~a convex combination of distribution functions.
This in turn is always a correctly normalized distribution.
In particular, when we compute the normalization constant, we see that 
\begin{equation}
\int \sumnode(\x) \, \mathrm{d}\x 
= 
\int
\sum_{\node \in \inn(\sumnode)} \theta_{\sumnode\node} \node(\x) \,\mathrm{d}\x 
= 
\sum_{\node \in \inn(\sumnode)} \theta_{\sumnode\node} \overbrace{\int \node(\x) \,\mathrm{d}\x}^{=1} 
= \sum_{\node \in \inn(\sumnode)} \theta_{\sumnode\node} = 1.
\end{equation}

Consequently, any node in a decomposable and smooth PC is a correctly normalized distribution over its scope, either by construction ($\distnode$), or because it is a factorized distribution ($\prodnode$) or a mixture distribution ($\sumnode$), whose factors (respectively mixture components) are given by other PC nodes.
We can figure that a PC structure is just a blueprint, crafting larger distributions out of smaller ones, like Lego blocks.

An important thing to note is that, even though products represent factorized distributions and hence independence between the scopes of their input nodes, the overall PC distribution does \emph{not} show these independence properties.
This fact stems from the mixture nodes, as a mixture of factorized distributions does (usually) not factorize.
One extreme example are Gaussian mixtures\footnote{Note that a Gaussian mixture models are just a special case of PC with a single sum node, whose input distributions are all Gaussian.} whose Gaussian components have diagonal covariance matrices and hence factorize into univariate Gaussians.
Even these models are known to be universal approximators of arbitrary densities, even though each of their components are individually assuming \emph{complete independence} among all random variables.

Decomposability and smoothness together with normalized sum weights yield a very nice interpretation of PCs as a \emph{hierarchical mixture model}.
However, an immediate question is whether we loose something when assuming normalized sum weights:
Are PCs with unnormalized weights more powerful, i.e., can we represent more distributions with it?
The answer to this question is negative \cite{peharz2015theoretical}:
Any decomposable and smooth PC with unnormalized weights can be converted into a PC with identical structure but normalized weights, while representing the exact same distribution.
As a warning, this procedure is not simply renormalizing the sum weights, at least not independently. 
Rather, it performs a bottom up sweep over the PC, and normalizes them in turn while compensating the scaling it does to higher nodes.
At any rate, however, this result gives us license to always assume that sum weights are normalized, without sacrificing any modeling power.

\section{Marginalization}

The most important consequence of decomposability and smoothness is that PCs allow to compute any marginal distribution in \emph{linear time} of the network size, i.e.~the number of edges $|E|$.
In order for this to hold true we require that the input distributions $\pmb{\distnode}$ are themselves tractable, i.e.~that they allow to compute any marginal.
A decomposable and smooth PC will basically ``inherit'' this tractability from its input distributions.

One way to ensure this is to use only one-dimensional distributions as input nodes, as marginalizing a single variable will simply deliver the constant $1$.
Another common choice is to use multivariate Gaussian, parametrized with a $D$-dimensional mean vector $\bm{\mu}$ and a $D \times D$ positive definite covariance matrix $\bm{C}$:
\begin{equation}
\label{eq:multivariate_Gauss}
p_{Gauss}(\x \cbar \bm{\mu}, \bm{C}) = 
{1 \over \sqrt{(2\pi)^D |\bm{C}|}}
\exp \left( -{1 \over 2} (\x - \bm{\mu})^T \bm{C}^{-1} (\x - \bm{\mu}) \right ).
\end{equation} 
It is a well-known fact that marginals of multivariate Gaussians are again Gaussian.
Moreover, these marginals are obtained by simply discarding the entries in $\bm{\mu}$ and $\bm{C}$ corresponding to marginalized random variables.

As we have seen in the last section, sum and product nodes in decomposable and smooth PCs are simply mixtures and factorized distributions of other distributions computed in the PCs.
The mechanism for tractable marginalization is that mixtures and factorizations allow to ``delegate'' marginalization to their mixture components and factors, respectively.

In particular, let $\sumnode$ be an arbitrary sum node in a decomposable and smooth PC, computing the mixture distribution
$$
\sumnode(\x) = \sum_{\node \in \inn(\sumnode)} \theta_{\sumnode\node} \node(\x).
$$
Assume we want to marginalize $\Z \subseteq \X$ and let $\Y = \X \setminus \Z$.
It holds that 
\begin{align*}
\sumnode(\y) 
& = \int \sum_{\node \in \inn(\sumnode)} \theta_{\sumnode\node} \, \node(\y, \z) \, \mathrm{d}\z \\
& =  \sum_{\node \in \inn(\sumnode)} \theta_{\sumnode\node} \int \node(\y, \z) \, \mathrm{d}\z \\
& =  \sum_{\node \in \inn(\sumnode)} \theta_{\sumnode\node} \, \node(\y), 
\end{align*}
meaning that the \textbf{marginal of a mixture is just the mixture of marginals}.

On the other hand, let $\prodnode$ be an arbitrary product node in a decomposable and smooth PC, computing the factorized distribution
$$
\prodnode(\x) = \node(\y) \, \node'(\z),
$$
where we assume that the product node has only two input nodes $\node,\node'$ (the case for arbitrarily many inputs works the same).
Due to decomposability, the scopes of the two nodes, $\Y = \X[\node]$ and $\Z = \X[\node']$, are disjoint.
Assume we want to marginalize a single dimension $X_i$ from $\prodnode$.
Due to decomposability, $X_i$ must appear either in $\Y$ or $\Z$, but not both.
Say it appears in $\Y$, then the marginal of $\prodnode$ is 
\begin{align*}
\int \prodnode(\x) \, \mathrm{d}x_i 
& = \int \node(\y) \, \node'(\z) \, \mathrm{d}x_i \\
& = \left( \int \node(\y) \, \mathrm{d}x_i \right) \, \node'(\z),  
\end{align*}
that is, $X_i$ gets just integrated out only from $\node$ since $\node'$ is constant with respect to $X_i$.
If we further want to integrate out another random variable $X_j$ which appears in $\Z$ we get
\begin{align*}
\int \int \prodnode(\x) \, \mathrm{d}x_i  \, \mathrm{d}x_j 
& = \int \int \node(\y) \, \node'(\z) \, \mathrm{d}x_i \, \mathrm{d}x_j \\
& = \left( \int \node(\y) \, \mathrm{d}x_i \right) \, \left( \int \node'(\z) \, \mathrm{d}x_j \right) ,  
\end{align*}
since $\node$ is now constant with respect to $X_j$.
Consequently, when we want to integrate out arbitrarily many random variables, the single-dimensional integrals just distribute to the factor where the corresponding variables appear, or in other words, the \textbf{marginal of a factorized distribution is just the factorization of the marginals}.

Thus, as long as the mixture components of sum nodes and the factors of product nodes ``know'' how they can do marginalization, the sum and product nodes just need to do their ``usual job''.
When we want to take a marginal of a PC, we apply the integrals to the root.
Since the root is either a sum or product node, we can reduce its marginal to the marginals of its input nodes, which are either distributions nodes or again sum and/or product nodes.
In the former case we are done, in the latter case we again reduce marginalization to the input nodes of the input nodes.
In that way, we recursively delegate the marginalization downwards until the input distributions are reached.
For these, however, we were assuming how to do marginalization in the first place.
We can summarize this principle by the definition of a marginal PC.

\begin{definition}[Marginal PC]
\label{def:marginal_pc}
Let a decomposable and smooth PC $\graph=(V,E)$ representing a joint distribution $p_{\X}$ be given, and let $\Y, \Z$ be a partition of $\X$, i.e., $\Y \cap \Z = \emptyset$, $\Y \cup \Z = \X$. 
Define another PC $\graph'=(V',E')$ which is an exact copy of $\graph$, except that each $\distnode \in V$ is replaced with a marginal distribution $\distnode' \in V'$, with $\Z \cap \X[\distnode]$ being marginalized out. 
In the case that $\Z \cap \X[\distnode] = \X[\distnode]$, i.e.~that all random variables are marginalized from the input distribution, $\distnode' \equiv 1$.
The PC $\graph'=(V',E')$ is called the $\Y$-marginal PC of $\graph$.
\end{definition}

From the considerations above it follows that the $\Y$-marginal PC computes the \textbf{exact} marginal $p_{\Y}$.
Note that the marginal PC might now contain constants $1$ as input nodes, so that  strictly speaking it is not a PC according to Definition~\ref{def:probabilistic_circuit}.
However, we might either interpret these constant input nodes as ``distribution over empty sets of random variables'' or we can prune them in linear time from the marginal PC, making it a proper PC according to Definition~\ref{def:probabilistic_circuit}.

\section{Conditioning}

With marginalization at hand, we also know how to compute conditionals, which are just a ratio between the joint and a marginal:
\begin{equation}
p(\y \cbar \z) 
= {p(\y, \z) \over p(\z)}
= {p(\y, \z) \over {\int p(\y, \z) \, \mathrm{d}\y}}
\end{equation} 
This requires that we compute two circuit evaluations, one for the joint and one for the marginal.
We can do better and construct a \emph{conditional PC} representing the conditional distribution explicitly.
The insight here is, like for marginalization, that sum and product nodes allow to delegate the conditioning operations to their input nodes.

Similarly as for marginalization, we require that conditioning is tractable for the input distributions.
This is easy for one-dimensional input distributions $\distnode(X)$, as the only possible conditional is to condition on the single variable delivering ${\distnode(x) \over \distnode(x)} = 1$, which is basically a distribution over an ``empty set of random variables,'' conditional on $X=x$.
For multivariate Gaussians \eqref{eq:multivariate_Gauss}, conditionals are available closed form.
Specifically, let a Gaussian over $\Y,\Z$ with parameters $\bm{\mu}$ and $\bm{C}$ be given.
The conditional over $\Y$ given $\Z=\z$ is again Gaussian with mean and covariance
\begin{equation}
\bm{\mu}_{\Y|\z} = \bm{\mu}_{\Y} + \bm{C}_{\Y\Z} \bm{C}_{\Z\Z}^{-1} (\z - \bm{\mu}_{\Z}) ~~~~~~~
\bm{C}_{\Y|\z} = \bm{C}_{\Y\Y} - \bm{C}_{\Y\Z} \bm{C}_{\Z\Z}^{-1} \bm{C}_{\Z\Y},
\end{equation}
where $\bm{\mu}_{\Y}$ and $\bm{\mu}_{\Z}$ are the subvectors of $\bm{\mu}$ containing the means corresponding to $\Y$ and $\Z$, respectively,  
$\bm{C}_{\Y\Y}$ and $\bm{C}_{\Z\Z}$ are the squared submatrices of $\bm{C}$ containing the (co-)variances corresponding to $\Y$, and $\Z$, respectively, and $\bm{C}_{\Y\Z}$ and $\bm{C}_{\Z\Y}$ are rectangular submatrices of $\bm{C}$, containing the covariances between $\Y$ and $\Z$.

Let $\sumnode$ be an arbitrary sum node in a decomposable and smooth PCs, computing the distribution
$$
\sumnode(\x) = \sum_{\node \in \inn(\sumnode)} \theta_{\sumnode\node} \node(\x),
$$
where without loss of generality we assume that $\X[\sumnode] = \X$.
Furthermore, also without loss of generality, we assume that the PC uses normalized sum weights and that each node is consequently a normalized distribution, as discussed in Section~\ref{sec:hierarchical_mixtures}

Assume we want to compute the conditional distribution $p_{\Y|\Z}$ where $\Y,\Z$ is some partition of $\X$.
We can use the fact that the conditional is proportional to the joint and write:
\begin{align}
\sumnode(\y \cbar \z) 
& \propto \sum_{\node \in \inn(\sumnode)} \theta_{\sumnode\node} \, \node(\y, \z) \\
& = \sum_{\node \in \inn(\sumnode)} \overbrace{\theta_{\sumnode\node} \, \node(\z)}^{\eqqcolon \tilde \theta_{\sumnode\node}} \, \node(\y \cbar \z) \\
& = \sum_{\node \in \inn(\sumnode)} \tilde \theta_{\sumnode\node}  \, \node(\y \cbar \z). 
\label{eq:sum_conditionals}
\end{align}
In \eqref{eq:sum_conditionals} we see that the conditional of $\sumnode$ is proportional to a linear combination of the conditionals of its inputs nodes.
The weights $ \tilde \theta_{\sumnode\node} = \theta_{\sumnode\node} \, \node(\z)$ are nonnegative but do not sum to one, while the conditionals  $\node(\y \cbar \z)$ are properly normalized distributions.
Hence, the proportionality factor must between \eqref{eq:sum_conditionals} and $\sumnode(\y \cbar \z)$ must be $\sum_{\node \in \inn(\sumnode)} {\tilde \theta_{\sumnode\node} }$, i.e.,
$$
\sumnode(\y \cbar \z) = \sum_{\node \in \inn(\sumnode)} {\theta_{\sumnode\node} \, \node(\z) \over \sum_{\node'} \theta_{\sumnode\node'} \, \node'(\z)} \, \node(\y \cbar \z),
$$
which means that the \textbf{the conditional of a mixture is a mixture of the conditionals}, with modified mixture weights $\bar \theta_{\sumnode\node'} \coloneqq {\theta_{\sumnode\node} \, \node(\z) \over \sum_{\node'} \theta_{\sumnode\node'} \, \node'(\z)}$.

On the other hand consider a product node
$$
\prodnode(\x) = \node(\x_1) \, \node'(\x_2),
$$
where we again assume that the product has only two inputs with scopes $\X_1$ and $\X_2$.
The more general case with more than two inputs is straightforward.
Let again $\Y,\Z$ be some partition of $\X$ and assume we want to compute
$\prodnode(\y \cbar \z)$.
Define $\Y_1 = \Y \cap \X_1$, $\Z_1 = \Z \cap \X_1$, $\Y_2 = \Y \cap \X_2$, $\Z_2 = \Z \cap \X_2$. 
The conditional is given as
$$
\prodnode(\y \cbar \z) = {\prodnode(\y,\z) \over \prodnode(\z)},
$$
which, since the marginal of a product is the product of the marginals, is
$$
\prodnode(\y \cbar \z) = {\node(\y_1, \z_1) \, \node'(\y_2, \z_2) \over \node(\z_1) \, \node'(\z_2)} 
= 
\node(\y_1 \cbar \z_1) \, \node'(\y_2 \cbar \z_2),
$$
i.e., \textbf{the conditional of a product is the product of the conditionals}.

Hence, similar as for marginalization, we can start at the root of a PC and ask the question what is its conditional distribution.
If it is a sum or product node, we have a simple recipe to express its conditional again as a sum and product node of the conditionals of its input distributions.
This ``delegation'' of the conditioning operator happens recursively until we reach the input distributions, for which we assumed tractable conditionals in the first place.
We can summarize these insights with the definition of the conditional PC.
\begin{definition}[Conditional PC]
\label{def:conditional_pc}
Let a decomposable and smooth PC $\graph=(V,E)$ representing a joint distribution $p_{\X}$ be given, and let $\Y, \Z$ be a partition of $\X$, i.e., $\Y \cap \Z = \emptyset$, $\Y \cup \Z = \X$.
Let a state $\z$ be given.
For each $\node \in V$, let $\node(\z)$ be the evaluation of the marginal PC, see Definition \ref{def:marginal_pc}.

Define another PC $\graph'=(V',E')$ which is an exact copy of $\graph$, except that each $\distnode \in V$ is replaced with its corresponding conditional distribution $\distnode'$ over $\Y \cap \X[\distnode]$, conditioned on $\z[\distnode]$.
In the case that $\Z \cap \X[\distnode] = \X[\distnode]$, i.e.~that all random variables are conditioned on, $\distnode' \equiv 1$.

Furthermore, replace for each sum node $\sumnode$ the weights as 
$$
\theta_{\sumnode\node} \leftarrow {\theta_{\sumnode\node} \, \node(\z) \over \sum_{\node'} \theta_{\sumnode\node'} \, \node'(\z)}
$$
The PC $\graph'=(V',E')$ is called the $\z$-conditional PC of $\graph$.
\end{definition}

Due to the considerations above, the $\z$-conditional PC represents the exact conditional $p_{\Y|\z}$ of the original PC.

\section{Expectations and Covariances}
\label{sec:expectations_covariances}

We see that decomposable and smooth PCs have excellent properties in terms of probabilistic inference, as the two core routines---marginalization and conditioning---can be be computed exactly and efficiently.
Furthermore, decomposability and smoothness also allow to compute the expectation (vector) and covariance (matrix) of a PC.

\paragraph{Expectation of PC.}
Following a similar argument as for marginalization, it can be shown that $\E[\X]$ can be computed exactly and efficiently when $p_{\X}$ is represented as a PC.
This follows from the fact that the expectation of a mixture is the mixture of expectations:
\begin{equation}
\label{eq:expectation_sum}
\E_{\sumnode}[\X] = 
\int 
\left(
\sum_{\node \in \inn(\sumnode)} \theta_{\sumnode\node} \, \node(\x)
\right) 
\, \x \, \mathrm{d}\x =
\sum_{\node \in \inn(\sumnode)} \theta_{\sumnode\node} \, 
\int
 \node(\x) \, \x \, 
\mathrm{d}\x
=\
\sum_{\node \in \inn(\sumnode)} \theta_{\sumnode\node} \, \E_{\node}[\X],
\end{equation}
and hence the expectation of a sum node is recursively given via the expectations of the sum's input nodes.

Furthermore, it is easily shown that the expectation of a factorized distribution is given as the concatenation of the expectations of the factors.
Hence, also the expectation of a product node is recursively given via the expectations of the product's input nodes.

This translates into a simple algorithm formulated as a network pass over the PC:
\begin{itemize}
\item For each input distribution $\distnode$, compute $\E_{\distnode}[\X[\distnode]]$, i.e.~the expectation of the distribution node over its scope.
\item For each sum node compute its expectation as in \eqref{eq:expectation_sum}.
\item For each product node, compute the expectation as the concatenation of the expectations of its input nodes.
\end{itemize}
In contrast to marginalization and conditioning, here the messages passed through the network are vector-valued.
The output of this procedure, the vector computed at the root node, will be exactly the expectation of the PC.

\paragraph{Covariances.}
Furthermore, the covariance of a PC can be computed efficiently, again via a recursive argument translating into a network pass.
Considering first a sum node computing the mixture
\begin{equation}
\label{eq:mixture_covariance}
\sumnode(\x) = 
\sum_{k = 1}^K \theta_{k} \, \node_k(\x),
\end{equation}
where $|\inn(\sumnode)| = K$ and we have enumerated the inputs to $\sumnode$ as $\node_1, \dots, \node_K$ using an arbitrary but fixed order.
A key technique in mixture models is to interpret them as \emph{latent variable models}.\footnote{This theme translates to PCs as well, as I have studied in \cite{peharz2016latent}.}
In particular, \eqref{eq:mixture_covariance} can be interpreted as the marginal distribution of an augmented model including a latent categorical variable $Z$ taking values in $\{1, \dots, K\}$
\begin{equation}
\label{eq:LV_covariance}
p_{\X,Z}(\x, k) = p_Z(k) \, p_{\X|Z}(\x \cbar k),
\end{equation}
where $p_Z(k) \coloneqq \theta_k$ and $p_{\X|Z}(\x \cbar k) \coloneqq \node_k(\x)$.
Indeed, marginalizing $Z$ from \eqref{eq:LV_covariance} yields exactly the original mixture \eqref{eq:mixture_covariance}.

The latent variable interpretation allows us to express the covariance $\mathbf{cov}[\X]$ via the \emph{law of total covariance}:
\begin{equation}
\label{eq:LV_total_covariance}
\mathbf{cov}[\X] = 
\E_Z \left[ \mathbf{cov}[\X \cbar Z] \right] + 
\mathbf{cov}_Z \left [ \E[\X \cbar Z]  \right ].
\end{equation}
Here $\mathbf{cov}[\X \cbar Z=k]$ is just the covariance of the \ith{k} input node to the sum node, which we assume to be tractable.
Hence the first term in \eqref{eq:LV_total_covariance} is simply the mixture of these covariances:
\begin{equation}
\E_Z \left[ \mathbf{cov}[\X \cbar Z] \right] = \sum_{k} \theta_k \, \mathbf{cov}[\X \cbar Z=k].
\end{equation}
Furthermore, $\E[\X \cbar Z=k]$ is just the expectation of the \ith{k} input node, which is computed as described in the previous paragraph.
By denoting $\bm{\mu}_k \coloneqq \E[\X \cbar Z=k]$ and the expectation of the sum node $\bm{\mu} \coloneqq \sum_{k} \theta_k \bm{\mu}_k$, we get that the second term in \eqref{eq:LV_total_covariance} is given as 
\begin{equation}
\mathbf{cov}_Z \left [ \E[\X \cbar Z]  \right ] 
=
\sum_{k} \theta_k (\bm{\mu}_k - \bm{\mu})^T (\bm{\mu}_k - \bm{\mu}).
\end{equation}
Hence, the covariance matrix of a sum node is recursively given via the covariances of its inputs, as in \eqref{eq:mixture_covariance}.

Furthermore, the covariance matrix of a factorized distribution is simply a block-diagonal matrix, whose on-diagonal blocks are given by the covariances of the factors and the remaining entries are zero (due to independence).
Here, without loss of generality, we assume that the random variables in the scopes of input nodes are ``neighboring'' in the covariance matrix---if they were not, the rows and columns would be permuted in some way.

In total, this again leads to an algorithm formulated as a network pass over the PC, where the messages passed through the network are matrix-valued.

\begin{itemize}
\item For distribution nodes compute covariance matrices directly from input distributions (e.g., covariance parameters of Gaussian nodes).
\item For sum nodes compute the covariance as in \eqref{eq:LV_total_covariance}.      
\item For product nodes construct a block-diagonal covariance matrix from input covariances:
\[
\mathbf{cov}[\X] =
\begin{pmatrix}
\mathbf{cov}[\X_1] & \mathbf{0} & \dots & \mathbf{0}\\[6pt]
\mathbf{0} & \mathbf{cov}[\X_2] & \dots & \mathbf{0}\\[6pt]
\vdots & \vdots & \ddots & \vdots\\[6pt]
\mathbf{0} & \mathbf{0} & \dots & \mathbf{cov}[\X_n]
\end{pmatrix}.
\]
\end{itemize}

\section{Determinism and Most Probable Explanation}

So far, we were assuming only decomposability and smoothness, which already brought us far.
Another powerful structural property is \emph{determinism}, which, in combination with decomposability, enables efficient computation of a \emph{most probable explanation} (MPE), i.e.~most likely joint assignment of variables:
\begin{equation}
MPE = \arg \max_x p(\x).
\end{equation}
Determinism in PCs is defined as follows.

\begin{definition}[Determinism]
A sum node $\sumnode$ in a PC is called \emph{deterministic} if at most one input node is non-zero for any given complete assignment $\x$:
\begin{equation}
\forall \x \colon |\{ \node \in \inn(\sumnode) : \node(\x) > 0 \}| \leq 1.
\end{equation}
A PC is deterministic of all its sum nodes are deterministic.
\end{definition}

Determinism and decomposability allow to compute an MPE with a single forward pass through the PC.
Like for the other inference routines above, this procedure follows by recursion.
First, consider a deterministic sum node $\sumnode$, for which we want to compute
\begin{equation}
\label{eq:sum_mpe_1}
\arg \max_{\x} ~
\sumnode(\x) = 
\arg \max_{\x} ~ \sum_{\node \in \inn(\sumnode)} \theta_{\sumnode\node} \, \node(\x).
\end{equation}
Due to determinism at most one input node is non-zero, for any $\x$, hence we can replace the summation in \eqref{eq:sum_mpe_1} by a maximum
\begin{equation}
\label{eq:sum_mpe_2}
\arg \max_{\x} ~
\sumnode(\x) = 
\arg \max_{\x} ~ \max_{\node \in \inn(\sumnode)} \theta_{\sumnode\node} \, \node(\x),
\end{equation}
which commutes with the maximum over $\x$:
\begin{align}
\label{eq:sum_mpe_2}
\arg \max_{\x} ~
\sumnode(\x) &= 
\arg \max_{\node \in \inn(\sumnode)} \max_{\x} ~  \theta_{\sumnode\node} \, \node(\x) \\
&= 
\arg \max_{\node \in \inn(\sumnode)}  \theta_{\sumnode\node} \, \max_{\x} ~ \node(\x). 
\end{align}
Thus, if we know how to compute an MPE for each input node $\node$, the MPE of $\sumnode$ is given by selecting the MPE of the input node that maximizes $\theta_{\sumnode\node} \, \node(\x)$.

Similarly, consider a decomposable product node
$
\prodnode(\x) = \prod_{\node \in \inn(\prodnode)} \node(\x).
$
It is easy to see that the product is maximized by maximizing individually each input node.
Hence the MPE of a decomposable product is simply the concatenation of the MPEs of its input nodes.
Consequently, MPE inference can be implemented as a simple forward pass through the PC, where messages are MPE solutions (vectors) together with the maxima they achieve:
\begin{itemize}
\item For each distribution node, compute the MPE assignment (which is assumed to be tractable).
\item For each sum node, select the MPE of the input node that maximizes the weighted density evaluated at its MPE.
\item For each product node, combine the MPE assignments from its inputs.
\end{itemize}

Determinism has other attractive properties as well.
In particular, it allows to compute a fast computation of (global) maximum likelihood parameters \cite{peharz2014learning} and a closed-form solution for Bayesian structure scores \cite{yang2023bayesian}.

\section{Further Structural Properties and Tractable Inferences}

We see that PCs achieve tractability by exploiting structural properties such as decomposability, smoothness, and determinism. 
Further structural constraints enable additional efficient inference routines and operations, extending the applicability of PCs to complex probabilistic queries \cite{vergari2021compositional}.

A strengthened notion of decomposability is \textit{structured decomposability}, which requires that two product nodes with identical scopes partition their scope identically.
Structured decomposability allows to construct tractable circuits representing operations such as products and quotients of circuits, powers of circuits, and logarithms of circuits, which are essential for ``advanced'' inference queries, in particular combinations of probability and logic.

Two structured-decomposable circuits with identical structure---but different parameters---are called \textit{compatible}, meaning they can be efficiently combined into a new structured-decomposable circuit without introducing exponential complexity. 
Some circuits are \textit{omni-compatibile}, which are compatible with \textit{any} other smooth and decomposable circuit over the same scope $\X$. 
Omni-compatibility often arises naturally, such as in fully-factorized models or mixtures thereof.
(Omni-)compatible PCs support various tractable inference routines beyond marginalization and conditioning, in particular exact computation of complex information-theoretic measures such as Kullback–Leibler divergence, cross-entropy, and mutual information.

See \cite{vergari2021compositional} for a systematic overview of these structural properties and the inferences they entail.

\section{Overview of Contributions}
\label{sec:contributions}

In the first part of this habilitation thesis, I highlight the role of \textbf{probability as an excellent core language for AI}: representations of joint distributions (probabilistic models) shall be interpreted as knowledge base, capturing both dependencies between random variables and their uncertainties.
This knowledge base is used for probabilistic reasoning, by combining various core inference routines, such as marginalization, conditioning, expectations and most probable explanation.

While probability constitutes a simple, elegant and optimal calculus of reasoning, the downside is that \textbf{probabilistic inference is a computational nightmare.}
Probabilistic circuits (PCs)---the central topic of this theses---are in a nutshell an elegant approach to probabilistic modeling as they ensure that \textbf{inference remains tractable}, that is, computable exactly in polynomial time.
The key to tractability lies in structural constraints, such as smoothness, decomposability, determinism, and structural decomposability, where different combinations of constraints lead to different tractable inference routines.

In conclusion of the first part, I provide a brief overview of 17 key papers I have published in the area of PCs in the last ten years, which are then provided in Part II.

\subsection{Foundations and Theory of PCs}

Part of my early work around PCs was focused on foundational aspects.
I was first exposed to these ideas via \emph{sum-product networks} (SPNs) \cite{poon2011sum},\footnote{Sum-product networks are the special case of smooth and decomposable PC.} which were proposed as a generalization of \emph{arithmetic circuits} (ACs) \cite{darwiche2003differential}.\footnote{Arithmetic circuits are basically equivalent to smooth, decomposable and deterministic PCs.}
The main proposed innovation in SPNs was that they were learned directly form data, that they did use \emph{unnormalized sum weights} and that they generalized the notion of decomposability to the notion of \emph{consistency}.
Thus, a natural question was how much these generalizations extend the modelling capabilities.

In \cite{peharz2015theoretical,peharz2015foundations} I have shown that they are actually quite vacuous: in particular, it turned out that any smooth and decomposable PC with unnormalized sum weights can be transformed into PC with the same structure and \emph{normalized} sum weights, still representing the same distribution.
This shows that using unnormalized weights does not improve the model power.
The transformation of sum weights can be done with a single sweep over the PC and hence is also quite efficient.
Further, I show in this paper that consistency is not significantly improving the model power either.
In particular, any smooth and consistent PC can be transformed into a smooth and decomposable PC by increasing the size of the PC at most by a factor linear in the number of random variables.\footnote{In order to show that one model class is significantly more expressive than another, one needs to show ``exponential separation'', i.e.~that there exists a family of distributions with polynomial representation size in the first model class and exponential size in the second.}

In \cite{peharz2016latent}, I addressed the interpretation of PCs as \emph{hierarchical latent variable models}, similar as finite mixture models (e.g.~Gaussian mixture models) can be interpreted as latent variable models with a single latent categorical variable.
This interpretation is important as it establishes a structured view on PCs, enabling the derivation of an exact ancestral sampling algorithm, the classical expectation-maximization algorithm, a sound interpretation of an approximate MPE algorithm proposed in \cite{poon2011sum}, and the interpretation of PCs as a Bayesian network (BN) \cite{koller2009probabilistic} involving latent variables.
The latter can be interpreted as a ``decompilation algorithm'', converting a PC into a BN, which we further studied in \cite{butz2020sum}.
The latent variable interpretation in PCs can further be exploited as a representation of data, as we have studied in \cite{vergari2018sum}.

Parts of the contributions in \cite{peharz2016latent} also appeared in \cite{peharz2015foundations}.
A further paper worth mentioning is with Trapp et al.~\cite{trapp2017safe}, where we proposed a safe semi-supervised learning algorithm for PCs, i.e.~using PCs as classifiers trained on both labelled and unlabelled data, with the guarantee that the performance on labelled data can only increase when adding unlabelled data.

\subsection{Bayesian Approaches with PCs}

Any parametric machine learning model can be rendered into a Bayesian version: rather than optimizing some loss function with respect to the model parameters one rather (i) equips the parameters with a prior distribution and (ii) infers a parameter posterior given data via Bayes rule.
Advantages of Bayesian approaches are that they are generally less prone to overfitting and represent model uncertainty explicitly.

In \cite{vergari2019automatic} we proposed a Bayesian inference algorithm for PC parameters based on Gibbs sampling.
Moreover, the leaves of PCs were generalized to mixtures over various exponential distributions representing \emph{different data types}, where inference over the data types was also included in the Gibbs sampler.
This, together with the latent variable interpretation, allowed to perform automatic recognition of the data type in an ``anonymous data table'' and further yielded highly-interpretable model, both by its structure and the tractable inference queries PCs facilitate.

In \cite{trapp2019bayesian} we used a similar Bayesian inference algorithm by Gibbs sampling, but extended the learning to \emph{PC structure}, allowing us to learn both parameters and structure in a single Bayesian setting.

A major drawback of Bayesian approaches in PCs is that they are, to a certain extent, in conflict with tractability: while smooth and decomposable PCs allow tractable marginalization and conditioning, the Bayesian version of them does not.
However, including the additional property of determinism and the use of factorized Dirichlet priors on the sum weights allows to perform Bayesian averages in closed form.
In \cite{yang2023bayesian} this property was used to derive \emph{Bayesian structure scores} \cite{heckerman1995learning} for deterministic PCs.
These scores are a principled score to compare different model structures in light of observed data and allow to learn an optimal structure in the infinite data limit.
While these scores were well-established in the graphical models literature \cite{koller2009probabilistic,heckerman1995learning}, our paper was---and to the time of writing this thesis still is---the only principled structure score proposed for PCs.

\subsection{Deep Learning and PCs}

Deep learning has without doubt been the most impactful technology in AI up to now.
While the field of deep learning us truly massive, its basic principle is rather simple: (i) express the learning task as a function approximation problem, (ii) construct an expressive function approximator via a (large) computational graph, i.e.~a composite function whose structure is described by directed acyclic graph over (differentiable) predefined functions, (iii) use automatic differentiation to compute gradients of model parameters, which is used in (iv) gradient-based optimization of a suitable loss function governed by the training data.
Many tricks of the trade contributed to the success of deep learning we are witnessing today, but a particular game changer was the development of dedicated libraries such as Tensorflow \cite{tensorflow2015-whitepaper} and PyTorch \cite{NEURIPS2019_9015}, which allow to implement and train deep models with ease.

PCs are also computational graphs and can be understood as special case of neural networks, whose input layer contain non-linear functions (densities) and whose higher layers either contain product or linear units.
Additionally, as discussed before, we apply some suitable subset of structural constraints in order to unlock tractable inference queries of our choice.

It is hence natural to use PCs as deep learning models as well.
In particular, they might be implemented in deep learning frameworks, benefit from GPU-acceleration and automatic differentiation, and might be easily interfaced with other deep learning models.
However, my first attempts to implement PCs on Tensorflow were naive, as I implemented each individual node as a separate node.
This led to huge and sparsely connected computational graph of small operations, which is a very GPU-unfriendly design.

The first decent implementation of PCs on Tensorflow were RAT-SPNs \cite{peharz2020random}, which vectorized input distributions and sum nodes, and product nodes were replaced with ``unrolled'' outer products.
The resulting computational graph had much fewer nodes and a higher degree of parallelism.
This implementation allowed to scale PCs to relatively large dataset such as (fashion)MNIST, but it was still about $50$ times slower than a comparable multi-layer perceptron (MLP).
The main reason for this bottleneck was the ``scattered'' and sparse graph induced by the PCs' structural constraints.

Taking lesson from these insights I developed \emph{einsum networks} (EiNets) a further accelerated implementation for PCs \cite{peharz2020einsum}.
The main idea in this paper was to further increase the degree of parallelism by ``summarizing'' several vectorized sum nodes into a bigger layer (tensor) and implementing all sum-product operations from one layer to the next with one monolithic call to the \emph{einsum}-function.\footnote{Einsum is implementing the Einstein sum-product contraction. It allows to represent all linear algebra operations as a parametrized call to einsum. It is readily implemented and optimized in deep learning frameworks.}
EiNets were now approximately en par with MLPs of comparable size,\footnote{My implementation, just using standard einsum, was just about $5$ times slower than MLPs. With further optimization, e.g. implementing dedicated GPU-kernels, this difference can be made even.}
and scaled to datasets sizes which were previously out of reach for PCs.
This central principle in this paper indicates a strong connection between PCs and tensor factorization, which has further been addressed in \cite{loconte2024relationship}.
Another interesting contributions in this paper was a simple implementation of EM which relied entirely on standard backpropagation.

As mentioned above, implementing PCs on deep learning frameworks makes them operationally compatible to neural networks.
In particular, one can now construct \emph{conditional PCs} \cite{shao2020conditional,shao2022conditional}, representing \emph{conditional distributions} $p(\y \cbar \x)$, by devising a PC structure over $\Y$, whose parameters $\bm{\theta}(\x)$ depend functionally on some context $\x$.
One is completely free how to represent $\bm{\theta}(\x)$ and might use arbitrary neural networks for this purpose.
As long as the PC satisfies the structural constraints with respect to $\Y$, it allows, for any $\x$, tractable inference in the distribution $p(\y \cbar \x)$.
Several of such conditional PC can also be combined into block-wise autoregressive models, for example as expressive generative models over images.

Furthermore, PCs lead to other interesting applications in deep learning.
In particular, if one allows for unconstrained (negative and positive) sum-weights and replaces the input densities with arbitrary non-linear functions, one yields a (non-probabilistic) circuit or \emph{decomposable neural network} \cite{subramaniexact}.
This architecture is less constrained than PCs and has interesting applications.
In particular, in \cite{subramaniexact}, we exploited the fact that expectations over diagonal (independent) Gaussians inputs can be tractably computed in deconets which can be used in exact robustness guarantees concerning adversarial attacks \cite{cohen2019certified}.

\subsection{Hybrid Models and Inference}

PCs make a fantastic promise, as probabilistic inference, which is generally NP-hard in most representations of joint distributions, is easy when the distribution is represented as a PC.
The catch is that this property weakens the expressive power of PCs.
One seems to be confronted with a dichotomy: 
Do we want tractable inference and hence go with PCs (or some other tractable model)?
Or do we need expressive power and hence go with highly expressive models such as VAEs, GANs, Flows, auto-regressive models, etc., and accept the fact that inference needs to be approximated?

It is suggestive that this is a \emph{false dichotomy} and that there actually exists an interesting (and largely unexplored) middle ground of \emph{hybrid models}: tractable where possible, approximate where required.
Some of my work has explored some options to achieve such hybrid models.

In \cite{stelzner2019faster} we have proposed a variation of the \emph{attend-infer-repeat} (AIR) system \cite{eslami2016attend} using \emph{PCs as a sub-module of an intractable model}.
Specifically, the AIR system is a method for Bayesian visual scene analysis, putting a prior on the scene, in particular on the number of objects, their category, scale, position, etc.
Further, it uses a rendering procedure to specify a likelihood of an observed image given a latent scene.
The goal is to infer the latent scene given an observed image, naturally formulated as Bayesian posterior inference.
The AIR system solves this task by using a recurrent neural network to infer objects sequentially and removing them from the scene.
In the generative model, the different object categories are represented with object-specific VAEs \cite{kingma2019introduction}. 
Hence, the entire model contains several thousand latent random variables to be inferred, in particular random variables describing the scene descriptions, pixels and VAE-codes of objects.
The inference method of AIR was entirely based on amortized variational inference.

In \cite{stelzner2019faster} we replaced the object models with PCs and devised a technique to marginalize them from the posterior, leaving just a few variables for the scene description to be inferred.
Hence, the entire system was not ``fully tractable'', but we solved large parts of the inference problem via tractable inference and inferred the rest with variational inference.
Here, the intuition is that intractable models over few variables are ``more tractable'' than intractable models over many variables.\footnote{Note that this is a rule-of-thumb line of reasoning and making this theoretically concrete requires much deeper thought.}
The main result in our paper was faster convergence of scene inference---even though we use the slow implementation of \cite{peharz2020random}---as well as more stable results of the variational inference routine.

Rather than taking PCs as sub-modules of intractable models one might also use \emph{intractable modes as sub-modules of PCs}.
Specifically, in \cite{tan2019hierarchical} we used VAEs \cite{kingma2019introduction} as PC leaves.
This is syntactically valid, as VAEs are merely infinite mixture distributions.
We further showed that stochastic ELBO estimates, which are used as learning signal in vanilla VAEs, can be used as replacement of log-likelihoods for the PC leaves, yielding an ELBO estimate for the hybrid PC/VAE model and hence a sound learning objective.
The main insight is, again, that the VAE leaves can be much smaller than a regular VAE (in particular their latent code can be of smaller dimensionality), since they model only part of the random variables.
Hence, the resulting model is a ``collection of many small intractable models, glued together via a tractable model'', which intuitively might yield a simpler inference problem.\footnote{Again, this is an qualitative argument.}
In our experiments we indeed showed that our hybrid model learned faster and achieved higher log-likelihoods than vanilla VAEs.

Basically the same idea was pursued in \cite{trapp2020deep} but using Gaussian processes (GPs) \cite{williams2006gaussian} as leaves rather than VAEs.
GPs actually do admit tractable inference, but require cubic time and quadratic memory in the number of samples, which becomes a notorious bottleneck for a large number of samples.
Besides using sparse inducing point techniques \cite{quinonero2005unifying}, the most popular technique to address this problem is to use a divide-and-conquer approach, by splitting the data points in subsets, performing GP inference on these subsets and aggregating the results.
In \cite{trapp2020deep}, we also followed this principle by devising a PC structure over local GP experts.
Unlike other divide-and-conquer approaches, our model was a sound and well-defined probabilistic model and still admitted \emph{exact} inference, resulting in more faithful inferences than the baselines.

In \cite{correia2023continuous} and \cite{gala2024probabilistic} we explored a further interesting principle to combine tractable models with intractable ones.
The basic idea in these papers was to augment the PC formalism with a new type of node: an integral node.
Formally, an integral node is equipped with a continuous latent random variable (or random vector) distributed according to some simple distribution, like Gaussian.
PC parameters below this integral node then depend functionally on this latent variable, yielding an infinite mixture of PCs.
Note that VAEs can also be described in this framework and just contain a single integral node.
Unlike VAEs, in \cite{correia2023continuous,gala2024probabilistic} we leveraged \emph{numerical integration} techniques (quadratures), for which approximation guarantees can be derived.
To this end, the dimensionality of the latent continuous variables need to be kept small enough, as numerical integration scales badly to large dimension.
The main outcome of these papers was that PCs with integral nodes outperform classical PCs.

\subsection{PCs and Symbolic Models}

PCs have a close connection to (tractable) logical circuits \cite{darwiche2002knowledge,darwiche2003differential}, which enables neuro-symbolic approaches by combining the two \cite{ahmed2022semantic}.
Moreover, they have natural connections to other machine learning models and symbolic methods.

In \cite{correia2020joints}, we established a connection between probabilistic circuits and decision trees.
In particular, decision trees can be transformed into PCs by transforming each decision node into a sum node over products of the sub-trees below the decision node and indicator nodes representing the respective decisions.
Additionally, 
\begin{itemize}
\item one attaches weights to the generated sum nodes given by the normalized counts of samples split by the decision, conditional on all decisions above the decision node; 
\item one also augments the classifier associated with each of the leaves---a conditional distribution over the prediction target---by multiplying it with a (simple) joint distribution over features using the samples associated to the leaf.
\end{itemize}
The resulting PC represents a joint distribution $p(\x,y)$, over both features $\x$ and target $y$ rather than a conditional model $p(y \cbar \x)$.
The conditional distribution derived from this joint, however, is under certain conditions exactly the same is the original decision tree classifier.  
Hence, learning a decision tree and transforming it into a PC can be seen as augmenting it into a full joint distribution, still yielding the same classifier in a ``backward compatible way''.
Advantages of this procedure are, among others, the possibility to treat missing input features by PC marginalization and to detect outliers by monitoring the probability of input features $p(\x)$.
Furthermore, also random forests can by augmented to generative models by re-interpreting them as mixtures of PCs obtained from the individual decision trees.

In \cite{loconte2023turn} we established a connection between \emph{knowledge graph embeddings} (KGEs) and PCs.
Knowledge graphs are a prominent symbolic representation of domain knowledge.
Formally, they are directed multi-graphs whose nodes are \emph{entities} and whose edges correspond to \emph{predicates} between entities.
The entity where an edge starts is called \emph{subject} and the entity where it ends is called \emph{object}.
The entire knowledge graph can be understood as a collection of subject-predicate-object (SPO) triplets. 
Typical tasks in knowledge graphs are \emph{link prediction} (which is important because most knowledge graphs are severely incomplete) and inferring new predicates, e.g.~by learning and applying rules among predicates

Knowledge graphs are typically used in machine learning applications by leveraging KGEs, obtained by constructing vector-valued representations for entities and predicates, which can be combined into an SPO embedding by applying some function to the S-, P- and O-embeddings.
A wide range of functions have been proposed in literature, where many of them are using some sum-product form, making them reminiscent to PCs.
In \cite{loconte2023turn}, we made this connection explicit and devised ways to modify common KGEs into bona-fide PCs.
The advantages of this approach are, among others, that standard maximum likelihood training can used to learn KGEs (instead of perhaps less principled techniques such as pseudo-likelihood and contrastive techniques), that our approach allows to sample from the distribution over SPO triplets, faster inference for triplets with missing entities or predicates and the possibility to encode hard logical constraints in KGEs.

Finally, in \cite{wedenig2024exact} we used logical circuits and probabilistic inference to construct the first exact approach for \emph{soft analytic side-channel attacks} (SASCA) \cite{veyrat2014soft}.
SASCA highlights vulnerability of cryptography algorithms (such AES) to physical information leaks, where any kind of physical side-channel might be used, such as power traces, heat dissipation, and hardware timing.
SASCA assumes that an attacker has access to a physical copy of a device running a cryptographic algorithm.
By querying the device with many random inputs of plain message and key, the attacker can learn so-called \emph{templates}, i.e.~probabilistic models mapping from side-channel to intermediate computations in the algorithm.
These ``soft guesses'' can be integrated by exploiting the logical structure of the cryptographic algorithm, which can be formulated as probabilistic inference in a factor graph \cite{loeliger2004introduction}, inferring the secret key from the observed side channel.

The standard approach in SASCA is to use \emph{loopy belief propagation} for inference, which has few theoretical guarantees and whose inference quality is hard to assess.
In \cite{wedenig2024exact}, we used a knowledge compilation approach \cite{darwiche2002knowledge} to compile large parts of the factor graph into a circuit representation and leveraged PC inference for \emph{exact inference} about the key.
The result was a dramatic increase of success rate and a higher tolerance against noise, which highlights further vulnerabilities against protected implementations of cryptographic algorithms.